\documentclass[10pt,twocolumn,letterpaper]{article}

\usepackage{cvpr}              

\usepackage[utf8]{inputenc} 
\usepackage[T1]{fontenc}    
\usepackage{url}            
\usepackage{booktabs}       
\usepackage{amsfonts}       
\usepackage{nicefrac}       
\usepackage{microtype}      
\usepackage{graphicx}
\usepackage{afterpage}
\usepackage{color}
\usepackage{wrapfig}
\usepackage{epsfig}
\usepackage{graphicx}
\usepackage{amsmath}
\usepackage{amssymb}
\usepackage{multirow}
\usepackage{algorithm}
\usepackage{arydshln}
\usepackage{threeparttable}
\usepackage{colortbl}
\usepackage{enumitem}
\usepackage{siunitx}
\usepackage{bm}
\usepackage{algpseudocode}
\usepackage[numbers]{natbib}
\bibpunct{[}{]}{,}{n}{,}{}
\usepackage{array}
\PassOptionsToPackage{table,x11names}{xcolor} 
\usepackage{xcolor} 
\usepackage{colortbl}
\usepackage{caption}
\usepackage{dblfloatfix}
\usepackage{amssymb}
\usepackage{pifont}
\usepackage{amssymb}
\usepackage{float}

\usepackage{listings}
\usepackage[export]{adjustbox}
\usepackage{makecell}
\usepackage{tikz}

\usepackage{subcaption}
\usepackage{array}
\usepackage{tabularx}
\definecolor{codeblue}{rgb}{0.25, 0.5, 0.5}
\definecolor{mygray}{gray}{.9}
\definecolor{ggray}{RGB}{127,127,127}
\definecolor{reda}{RGB}{192,0,0}
\definecolor{redb}{RGB}{217,148,143}
\definecolor{myyellow}{RGB}{190,144,0}
\definecolor{mygreen}{RGB}{80,100,40}
\definecolor{myblue}{RGB}{30,90,100}
\definecolor{brightgreen}{rgb}{0.0, 1.0, 0.0} 
\definecolor{brightred}{rgb}{1.0, 0.0, 0.0}

\definecolor{MatplotlibRed}{rgb}{1, 0, 0}        
\definecolor{MatplotlibBlue}{rgb}{0, 0, 1}       
\definecolor{MatplotlibOrange}{rgb}{1, 0.5, 0}   
\definecolor{MatplotlibPurple}{rgb}{0.5, 0, 0.5} 
\definecolor{MatplotlibCyan}{rgb}{0, 1, 1}       
\definecolor{MatplotlibGreen}{rgb}{0, 0.5, 0}    

\newcommand{\reshl}[2]{
\textbf{\fontsize{9pt}{1em}\selectfont #1}\fontsize{7.5pt}{1em}\selectfont{\color{brightgreen}{$\uparrow$\textbf{#2}}}
}
\newcommand{\reshldown}[2]{
\textbf{\fontsize{9pt}{1em}\selectfont #1}\fontsize{7.5pt}{1em}\selectfont{\color{brightgreen}{$\downarrow$\textbf{#2}}}
}
\newcommand{\reshldownred}[2]{
\textbf{\fontsize{9pt}{1em}\selectfont #1}\fontsize{7.5pt}{1em}\selectfont{\color{brightred}{$\downarrow$\textbf{#2}}}
}
\newcommand{\reshlup}[2]{
\textbf{\fontsize{9pt}{1em}\selectfont #1}\fontsize{7.5pt}{1em}\selectfont{\color{brightred}{$\uparrow$\textbf{#2}}}
}

\newcommand{\pub}[1]{{\color{gray}{\tiny{[{#1}]\!}}}}

\newcolumntype{x}[1]{>{\centering\arraybackslash}p{#1pt}}

\makeatletter\renewcommand\paragraph{\@startsection{paragraph}{4}{\z@}
    {.5em \@plus1ex \@minus.2ex}{-.5em}{\normalfont\normalsize\bfseries}}\makeatother

\usepackage{siunitx}
\usepackage[nopar]{lipsum}
\usepackage[export]{adjustbox}

\makeatletter
\newcommand{\thickhline}{%
    \noalign {\ifnum 0=`}\fi \hrule height 1pt
    \futurelet \reserved@a \@xhline
}

\makeatother

\usepackage{caption}
\usepackage[utf8]{inputenc}

\definecolor{bblue}{RGB}{0,30,95}
\definecolor{rred}{RGB}{190,0,0}
\definecolor{mygray}{gray}{0.1}
\definecolor{ggray}{RGB}{127,127,127}
\usepackage[pagebackref=true,breaklinks=true,letterpaper=true,colorlinks,bookmarks=false]{hyperref}

\definecolor{cvprblue}{rgb}{0.21,0.49,0.74}

\def\paperID{*****} 
\def\confName{CVPR}
\def\confYear{2025}

\title{Balanced Gradient Sample Retrieval for Enhanced Knowledge Retention in Proxy-based Continual Learning}

\author{
Hongye Xu\thanks{Equal contribution.}\\
Rochester Inst. of Tech.\\
{\tt\small hx5239@rit.edu}
\and
Jan Wasilewski\footnotemark[1]\\ 
Rochester Inst. of Tech.\\
{\tt\small jw7630@rit.edu }
\and
Bartosz Krawczyk\thanks{Corresponding author.}\\
Rochester Inst. of Tech.\\
{\tt\small Bartosz.Krawczyk@rit.edu}
}

\begin{document}
\maketitle
\begin{abstract}
Continual learning in deep neural networks often suffers from catastrophic forgetting, where representations for previous tasks are overwritten during subsequent training. We propose a novel sample retrieval strategy from the memory buffer that leverages both gradient-conflicting and gradient-aligned samples to effectively retain knowledge about past tasks within a supervised contrastive learning framework. Gradient-conflicting samples are selected for their potential to reduce interference by re-aligning gradients, thereby preserving past task knowledge. Meanwhile, gradient-aligned samples are incorporated to reinforce stable, shared representations across tasks. By balancing gradient correction from conflicting samples with alignment reinforcement from aligned ones, our approach increases the diversity among retrieved instances and achieves superior alignment in parameter space, significantly enhancing knowledge retention and mitigating proxy drift. Empirical results demonstrate that using both sample types outperforms methods relying solely on one sample type or random retrieval. Experiments on popular continual learning benchmarks in computer vision validate our method’s state-of-the-art performance in mitigating forgetting while maintaining competitive accuracy on new tasks.
\end{abstract}    
\section{Related Work}
\label{in}
\vspace{-6pt}

Continual learning, the capability of a machine learning model to sequentially learn from a stream of data while retaining knowledge from previous tasks, is essential for the development of intelligent systems that can adapt and evolve over time \cite{Parisi:2019}. However, traditional machine learning algorithms often struggle with continual learning scenarios due to a phenomenon known as catastrophic forgetting, wherein the model rapidly loses performance on previously learned tasks as it learns new ones \cite{Lange:2022}. This challenge poses a significant obstacle to the deployment of lifelong learning systems in real-world applications.\cite{zenke2017continual}

To address this issue, various strategies~\cite{zenke2017continual,lopez2017gradient,schwarz2018progress,aljundi2018memory,riemer2018learning,serra2018overcoming,saha2020gradient,pham2021dualnet,caccia2021new,deng2021flattening,cha2021co2l,wang2022coscl,wang2022meta,slim2022_transil,wang2023dualhsic,yang2023efficient,UDIL,wang2024a} have been explored, among which memory-based methods~\cite{lopez2017gradient,riemer2018learning,pham2021dualnet,caccia2021new,wang2022meta,wang2023dualhsic,yang2023efficient,UDIL,wang2024a} have emerged as promising solutions. Memory-based approaches leverage the idea of storing past experiences in a memory buffer and utilizing them to mitigate forgetting when learning new tasks \cite{Buzzega:2020}. These methods range from simple replay buffers that randomly sample and replay past data, to more sophisticated memory architectures equipped with mechanisms for selective rehearsal, regularization, and transfer of knowledge~\cite{mallya2018packnet,nguyen2018variational,hung2019compacting,li2019learn,kurle2019continual,titsias2019functional,cha2020cpr,pan2020continual,kao2021natural,henning2021posterior,rudner2022continual}. 


Recently, contrastive learning emerged as a promising approach to continual learning by allowing for structuring feature spaces that are robust to task shifts and less prone to catastrophic forgetting \cite{lin2023pcr}. Unlike cross-entropy-based methods, which rely on static class boundaries and task-specific memory, contrastive learning optimizes for relational alignment between samples, enabling better feature transfer across tasks. However, these methods suffer from the drift phenomena that continuously reduces the relevance of proxies (prototypes) and increases catastrophic forgetting \cite{Korycki:2021,Gomez:2024,Goswami:2024}. Additionally, existing memory-based approaches designed for cross-entropy loss struggle in the contrastive setting due to fundamental differences in how representations are maintained. 

\noindent \textbf{Research goal.} To propose strategy for targeted sample retrieval from the memory buffer designed for supervised contrastive continual learning that balances maintaining stable, shared representations among tasks and preserving past knowledge to mitigate catastrophic forgetting and proxy drift. 

\noindent \textbf{Motivation.} While contrastive loss can be seen as a very attractive approach to continual learning, the existing memory-based algorithms were designed with cross-entropy loss in mind and thus do not translate well to this setup. First, cross-entropy-based methods focus on preserving class boundaries, whereas contrastive learning relies on relational stability between positive and negative pairs, making traditional memory replay or regularization approaches ineffective. Second, contrastive continual learning requires innovative mechanisms to prevent proxy drift across tasks and to preserve pairwise relationships. Cross-entropy-based methods fail to capture the nuanced relational structures needed for contrastive-based tasks, leading to rapid forgetting and biased representations.



\noindent \textbf{Main contributions.} This work offers the following contributions to the continual learning domain:
\begin{itemize}[leftmargin=*]
\item New approach to instance retrieval from the buffer for memory-based contrastive continual learning based on combination of gradient-aligned and gradient-conflicting samples.
\item Theoretical analysis of the proposed solution showing that gradient-aligned instances are useful for finding stable embeddings shared among tasks, while gradient-conflicting ones are useful for preserving past knowledge and avoiding catastrophic forgetting.
\item Experimental study comparing the proposed approach with state-of-the-art continual learning algorithms and buffer instance selection approaches for memory-based methods, showing that our methods lead to increased diversity of retrieved instances from the buffer, significantly reduced proxy drift, and superior accuracy and knowledge retention.  
\end{itemize}

\section{Related Work}
\label{rw}
\vspace{-6pt}

\noindent\textbf{Continual Learning.} This domain addresses the fundamental challenge of enabling neural networks to acquire new knowledge sequentially while maintaining previously learned information. While McCloskey and Cohen's seminal work~\cite{mccloskey1989catastrophic} identified catastrophic forgetting as a critical limitation of neural networks. Since then, numerous researchers have addressed this issue. For instance, Li and Hoiem \cite{LWF} introduced Learning without Forgetting (LwF), which uses knowledge distillation to preserve previous task performance without storing old data. Kirkpatrick et al.~\cite{kirkpatrick2017overcoming} proposed the Elastic Weight Consolidation (EWC), which estimates parameter importance and selectively constrains weight updates to prevent catastrophic forgetting. PackNet \cite{PackNet} introduces an innovative approach by iteratively identifying and freeing redundant network parameters for new tasks. Progressive Neural Networks~\cite{Fayek:2020, ProgresiveNNJJ} maintain separate networks for each task while establishing lateral connections to leverage previous knowledge.

\noindent\textbf{Memory based methods.} Another notable approach includes memory-based methods such as Experience Replay \cite{Rolnick:2019} with multiple variations \cite{Boschini:2023, weightInterpolationJJ, chaudhry2018efficient} that store and revisit old data to prevent forgetting. Due to memory constraints, storing all data is impractical, so effective buffer composition is key. Recent studies suggest that strategic instance selection is crucial in creating an effective memory for continual learning~\cite{Krutsylo:2024}. For example, bilevel coreset selection uses bilevel optimization to identify compact and representative sample subsets, ensuring that they adequately capture past knowledge without overwhelming the buffer~\cite{Hao:2023}. Krutsylo and Morawiecki’s diverse memory framework enhances this by choosing varied examples, reducing redundancy and broadening generalization~\cite{diversity:2022}. Information-theoretic memory selection maximizes information gain, carefully selecting high-value samples that promote efficient memory use~\cite{Sun:2022}. Similarly, entropy-based sample selection targets high-uncertainty samples, focusing learning on challenging cases that reinforce robustness~\cite{wiewel2021}.
 
\noindent\textbf{Contrastive Learning in Continual Learning.} Contrastive learning has recently been applied to continual learning to mitigate the problem of catastrophic forgetting. Recent works have explored various strategies to leverage contrastive learning for effective knowledge preservation. Supervised Contrastive Replay (SCR) \cite{SCR} and Proxy-based Contrastive Replay (PCR) \cite{lin2023pcr} demonstrate effective strategies for maintaining feature representations across tasks. While SCR focuses on preserving semantic relationships through contrastive memory replay, PCR introduces class proxies to reduce memory requirements and enhance discrimination between classes from different tasks. CO2L \cite{Co2L} further advances this field by using contrastive learning to align feature spaces between old and new tasks.

\noindent\textbf{Targeted retrieval.} Another line of work investigates techniques for optimizing sample selection from the experience replay buffer. Diversity sampling chooses a range of examples to avoid overfitting, maintaining knowledge variety~\cite{diversity:2022}. Another method, ASER (Adversarial Shapley Value Experience Replay) leverages Shapley values to identify samples that significantly impact performance, balancing those that aid and challenge the model~\cite{shim2021online}. MIR (Maximally Interfering Retrieval) selects examples that cause high parameter updates if retrained, focusing on those prone to forgetting~\cite{aljundi2019online}. SWIL (Selective Weight Importance Learning) targets samples that engage essential weights from prior tasks, preserving critical knowledge~\cite{SWIL}. GRASP (Gradient-Sensitive Prioritization) uses gradient-based metrics to choose samples that stabilize past knowledge, creating a buffer that effectively balances stability and adaptability~\cite{GRASP}.
\section{Methodology}\label{Method}
\label{Method}
In this section, we introduce our approach to improving continual learning performance by strategically retrieving positive and negative loss change samples within a contrastive learning framework that utilizes proxies. We first formulate the continual learning problem under experience replay, discuss the proxy drift problem, outline our sample retrieval strategy, and provide a theoretical analysis demonstrating the superiority of our method in mitigating catastrophic forgetting and proxy drift, as well as promoting effective task learning.

\subsection{Problem formulation}
In continual learning, a model is exposed to a sequence of tasks, denoted as \( \mathcal{T} = \{T_1, T_2, \dots, T_n\} \), where each task \( T_i \) is associated with a dataset \( D_i = \{(x_i^j, y_i^j)\}_{j=1}^{N_i} \) containing input-output pairs \( (x_i, y_i) \) sampled from a task-specific distribution \( P_i(X, Y) \).

The goal is to train a model \( f_\Phi \) parameterized by $\Phi$, which can learn each task \( T_i \) sequentially, while retaining knowledge of previously learned tasks without re-accessing their data (i.e., avoiding catastrophic forgetting). Formally, we aim to find an optimal parameter set $\Phi$ that minimizes the cumulative loss over all tasks seen so far:

\begin{equation}
\Phi^* = \arg \min_\Phi \sum_{i=1}^n \mathbb{E}_{(x, y) \sim P_i} \left[ \mathcal{L}(f_\theta(x), y) \right],
\end{equation}

where \( \mathcal{L} \) is the loss function (e.g., cross-entropy loss for classification).


\subsection{Proxy-Based Contrastive Learning}
In our method, we integrate proxy-based contrastive learning.  Let $\Phi_h$ be the feature extractor parameters, $h(x; \Phi_h)$ be the embedding of input $x$, and $w_y$ be the trainable proxy for class $y$. The training objective encourages $h(x; \Phi)$ to align with its corresponding proxy $w_y$ while diverging from other class proxies.

The proxy-based contrastive loss is defined as:

\begin{equation}
\mathcal{L}_{\text{ctr.}} = \mathbb{E}_{(x, y) \sim \mathcal{D}}  -\log \left( \frac{\exp\left( \gamma \langle h(x; \Phi_h), w_y \rangle \right)}{\sum_{c \in \mathcal{C}^{(t)} \cup \mathcal{C}^{(\text{buffer})}} \exp\left( \gamma \langle h(x; \Phi_h), w_c \rangle \right)} \right) ,
\label{eq:contrastive_loss}
\end{equation}

where $\gamma$ is a scaling factor, $\langle \cdot, \cdot \rangle$ denotes the cosine similarity, and $\mathcal{C}^{(t)} \cup \mathcal{C}^{(\text{buffer})}$ is the set of class indices in the current training batch and classes of the samples selected from the buffer.

\subsection{Continual Learning with Experience Replay}\label{PF}
To mitigate catastrophic forgetting the tendency of the model to lose previously learned knowledge when updated with new task data—a memory buffer \( \mathcal{M} \) is maintained. This buffer stores a limited number of representative samples from previous tasks, allowing the model to rehearse prior knowledge as new tasks are introduced. Formally, let \( \mathcal{M} = \bigcup_{i=1}^{n} \mathcal{M}_i \), where each \( \mathcal{M}_i \subset D_i \) is a subset of samples from task \( T_i \) with size \( |\mathcal{M}| = M \), and \( M \) is the memory budget.

The training objective is to minimize the combined loss over the current task \( T_i \) and the memory buffer \( \mathcal{M} \):

\begin{align}
\Phi^* = \arg \min_\Phi \Big( & \, \mathbb{E}_{(x, y) \sim P_i} 
\left[ \mathcal{L}_{ctr.}(f_\theta(x), y) \right] \nonumber \\
& +   \mathbb{E}_{(x, y) \in \mathcal{M}} 
\left[ \mathcal{L}_{ctr.}(f_\theta(x), y) \right] \nonumber \Big).
\end{align}
Samples are added to the memory buffer by reservoir sampling, the method maintaining a uniform sampling of data across all tasks.

\subsubsection{MIR method}
Our work focuses on samples selection from the buffer. In particular, we build on The Maximum Interference Retrieval~\cite{aljundi2019online}, the method that strategically selects memory samples for replay based on their vulnerability to interference. Instead of replaying random or recent samples, MIR identifies those samples in the memory buffer expected to experience the highest increase in loss, or maximum interference, if the current task's update were applied. By simulating gradient updates, MIR estimates this interference for each sample, prioritizing those that show the greatest potential loss increase. These high-interference samples are then replayed alongside new task samples during training, ensuring that the model retains essential knowledge from previous tasks without excessive memory usage. This selective rehearsal reduces catastrophic forgetting by focusing learning on the most at-risk past samples, maintaining accuracy across tasks. The method was originally proposed for online continual learning, but we find that it also performs very well in standard continual learning scenario. To facilitate clarity in our discussion, we define a sampling method that is the opposite of MIR, which selects samples from the memory buffer anticipated to show the highest decrease in loss or minimal interference. We refer to this method as inverse-MIR (\textbf{IMIR}). This term will be used in the subsequent analysis for better understanding and consistency.

\subsubsection{Gradient alignment of selected samples and samples from the batch}
In this section we analyse MIR algorithm. We denote a cross-entropy loss for sample $x$ as $l_{ce}(x, \Phi)$, and $\mathcal{L}_{ce}(X_t, \Phi)$ to be loss over the batch of samples $X_t = \{x_{i}^t\}_{i=1}^N$ 
\begin{equation}
    \mathcal{L}_{ce}(X_t, \Phi) = \frac{1}{N} \sum_{i=1}^{N} l_{ce}(x_i, \Phi).
\end{equation}
We define criterion $\mathcal{C}$:
\begin{equation}
    \mathcal{C}(x) = l_{ce}(x, \Phi') - l_{ce}(x, \Phi),
\end{equation}
We expand sample loss into first order Taylor series:
\begin{equation}
    l_{ce}(x, \Phi')  \approx l_{ce}(x, \Phi) + (\Phi' - \Phi)^\top \nabla_{\Phi} l_{ce}(x, \Phi).
\end{equation}
Since
\begin{equation}
\Phi' = \Phi - \alpha \nabla_\Phi \mathcal{L}_{ce} (X_t, \Phi),
\end{equation}
we can approximate $\mathcal{C}$ by:
\begin{equation}
    \mathcal{C}(x) \approx  - \alpha (\nabla_\Phi \mathcal{L}_{ce} (X_t, \Phi))^\top \nabla_{\Phi} l_{ce}(x, \Phi).
\end{equation}
Therefore, samples with the highest $\mathcal{C}$ have the most misaligned gradients with the current task gradient. Including them in the training process improves knowledge retention but limits model plasticity.


\subsection{Proxy Drift}


Contrastive continual learning is known to suffer from a phenomenon known as proxy drift, where class representations (proxies or prototypes) shift unpredictably as new tasks are introduced, due to distribution shift \cite{Korycki:2021,Gomez:2024,Goswami:2024}. This drift destabilizes learned representations and leads to increased catastrophic forgetting.

Proxy (or any prototype-based) drift for class $c$ between tasks $t-1$ and $t$ can be quantified by calculating the distance between $p_c^{(t)}$ and $p_c^{(t-1)}$ \cite{Gomez:2024}:
\vspace{-2mm}
\begin{equation}
\Delta p_c^{(t)} = \|p_c^{(t)} - p_c^{(t-1)}\|_2,
\end{equation}
\noindent where $\|\cdot\|_2$ denotes the Euclidean norm. A larger $\Delta p_c^{(t)}$ indicates greater drift for the proxy of class $c$, suggesting it has shifted significantly. 

The maintained memory has a crucial role in the proxy update. Thus to mitigate the proxy drift, experience replay with diverse data retrieval is critical, as it maintains a stable proxy space, aligning class representations to past knowledge while enabling effective adaptation to new tasks. We argue that existing instance selection methods rely on maximization of a single criterion and thus select samples with reduced diversity (highly concentrated around single preferred solution). The sampling method proposed in this paper combines gradient-aligned and gradient-conflicting instances, thus boosting the diversity of retrieved instances and offering better representation of past task distributions. 


\subsection{Balanced Retrieval}
We propose a sample retrieval strategy based on MIR sample selection criterion \cite{aljundi2019online}. We calculate the change of the contrastive loss with proxies from all previous classes in denominator, which is equivalent to cross-entropy loss used in original paper
\begin{equation}
l_{\text{ce}}(x,y) = -\log \left( \frac{\exp\left( \gamma \langle h(x; \Phi), w_y \rangle \right)}{\sum_{c \in \mathcal{C}^{(t)} \cup \mathcal{C}^{(\text{past})}} \exp\left( \gamma \langle h(x; \Phi), w_c \rangle \right)} \right),
\label{eq:contrastive_loss}
\end{equation}
In contrast to MIR, we select samples from \textbf{both} ends of loss-change distribution from the memory buffer. By selecting samples with the highest loss change we ensure knowledge retention and by selecting samples with the lowest loss change we reinforce model plasticity. This strategy aims to balance the contributions of the gradient during training, taking advantage of the complementary benefits of both types of samples. The comprehensive pseudo-code for our balanced sampling algorithm is detailed in Algorithm \ref{alg:modified_mir}.

\begin{figure}[htbp]
\begin{minipage}{0.95\columnwidth} 
\vspace{-4mm} 
\begin{algorithm}[H]
\caption{Balanced Retrieval}
\label{alg:modified_mir}
\begin{algorithmic}[0]
\Require Model $h$, Sample size $2C$, Task Data $D_t$, Memory $\mathcal{M}$, Parameters $\Phi$
\For{$t = 1$ \textbf{to} $T$}
    \For{each $X_n$ \textbf{in} $D_t$}
        \State \textcolor{codeblue}{\%\ Would-be parameter estimate}
        \State $\Phi' \gets \text{Optimizer}(h, X_n, \Phi, \alpha)$
        \State \textcolor{codeblue}{\% Select $C$ samples}
        \State $X_{C}^1 \gets \text{RandomSampling}(\mathcal{M}, C)$
        \State \textcolor{codeblue}{\% Descending sort and select top $n_1$ samples}
        \State $\{S_i^1\}_{i=1}^{n_1} \gets \text{Sort}(LossChange{}(X_{C}^1, \Phi, \Phi'))$
        \State \textcolor{codeblue}{\% Select another $C$ samples}
        \State $X_{C}^2 \gets \text{RandomSampling}(\mathcal{M}, C)$
        \State \textcolor{codeblue}{\% Ascending sort and select top $n_2$ samples}
        \State $\{S_i^2\}_{i=1}^{n_2} \gets \text{Sort}(LossChange{}(X_{C}^2, \Phi, \Phi'))$
        \State \textcolor{codeblue}{\% Combine the selected samples}
        \State $X_{\mathcal{M}_C} \gets \{S_i^1\}_{i=1}^{n_1} \cup \{S_i^2\}_{i=1}^{n_2}$
        \State $\Phi \gets \text{Optimizer}(X_n \cup X_{\mathcal{M}_C}, \alpha)$
        \State \textcolor{codeblue}{\% Add samples to memory}
        \State $\mathcal{M} \gets \text{UpdateMemory}(X_n)$
    \EndFor
\EndFor
\end{algorithmic}
\end{algorithm}
\end{minipage}
\vspace{-2mm} 
\end{figure}

\section{Experiments}

\label{Exp}
\subsection{Experimental Setup}\label{ES}

\textbf{Datasets.} We test our method with 6 datasets(\textit{CIFAR100 \cite{krizhevsky2009learning}}, \textit{Core50 \cite{lomonaco2017core50}}, \textit{Food100} is a subset of Food-101~\cite{bossard2014food}, \textit{Mini-ImageNet \cite{vinyals2016matching}}, \textit{Places100\cite{zhou2017places}} and \textit{Tiny-ImageNet \cite{le2015tiny}}). They were selected to include different types of visual content and thus test the efficacy of our proposed technique.

\noindent\textbf{Training.} Our training framework is inspired by the methodologies used in Online-Continual-Learning \cite{shim2021online} and Mammoth \cite{Boschini:2023,buzzega2020dark}, following their standard training protocols for all baseline algorithms. We adopt ResNet18~\cite{he2016deep} as the backbone, initializing all layers from scratch. Training is performed using Stochastic Gradient Descent (SGD) with an initial learning rate of 0.1. The training setup utilizes a batch size of 10, with an equal number of samples drawn from the buffer for each batch. This approach maximizes learning efficiency and effectively balances memory usage.

\noindent\textbf{Data augmentations.} Consistent data augmentation is applied across all datasets to enrich diversity and improve generalization. This includes random cropping to varying square dimensions, horizontal flipping, and color jittering that adjusts brightness, contrast, saturation, and hue with an 80\% probability. To further diversify the training data, images have a 20\% chance of being converted to grayscale.

\noindent\textbf{Spliting data into task streams.} Datasets are structured into tasks with distinct categories: CIFAR-100, Food-100, Mini-ImageNet, and Places-100 are divided into 20 tasks (each task comprising 5 classes), while Tiny-ImageNet is divided into 20 tasks with 10 classes each. Core50 is organized into 10 tasks, each containing 5 classes. Each task is trained sequentially for 50 epochs to ensure comprehensive learning.

\noindent \textbf{Testing.} All experimental results are presented using a test batch size of 256. To ensure fairness, \textit{no test-time data augmentation is applied}.

\begin{table*}[t]
	\centering
	\caption{{Quantitative results} on CIFAR100~\cite{krizhevsky2009learning},   Core50~\cite{lomonaco2017core50}, Food100, Mini-ImageNet~\cite{vinyals2016matching}, Places100 and Tiny-ImageNet~\cite{le2015tiny}. See \S\ref{label:re} for details.}
	\setlength\tabcolsep{6pt}
	\renewcommand\arraystretch{1.1}
	\resizebox{\linewidth}{!}{
		\begin{tabular}{c|cc|cc|cc|cc|cc|cc}
			\thickhline
			Datasets $\&$ & \multicolumn{2}{c|}{CIFAR100}&\multicolumn{2}{c|}{Core50} &\multicolumn{2}{c|}{Food100} &\multicolumn{2}{c|}{Mini-ImageNet}&\multicolumn{2}{c|}{Places100}&\multicolumn{2}{c}{Tiny-ImageNet}\\
                Method & \textit{Acc. ($\uparrow$ )} &\textit{Fgt. ($\downarrow$ )} &\textit{Acc. ($\uparrow$ )} &\textit{Fgt. ($\downarrow$ )} &\textit{Acc. ($\uparrow$ )} &\textit{Fgt. ($\downarrow$ )} &\textit{Acc. ($\uparrow$ )} &\textit{Fgt. ($\downarrow$ )} &\textit{Acc. ($\uparrow$ )} &\textit{Fgt. ($\downarrow$ )} &\textit{Acc. ($\uparrow$ )} &\textit{Fgt. ($\downarrow$ )}\\
			\hline
            ER\pub{ICMLW2019}\cite{chaudhry2019tiny} & 24.12 & 40.47 & 36.15 & 19.36 & 16.94 & 35.00 & 11.86 & 37.21 & 21.54 & 38.02 & 15.60 & 42.21\\ 
            A-GEM\pub{ICLR2019}\cite{chaudhry2018efficient} & 4.53 & 74.91 & 10.94 & 62.78 & 4.60 & 67.13 & 4.28 & 73.89 & 4.13& 72.52 & 3.43 & 70.12\\
            DER++\pub{NeurIPS2020}\cite{buzzega2020dark} & 21.50 & 41.08 & 53.65 & 10.22 & 12.54 & 33.92 & 18.52 & 42.03 & 17.33 & 38.01 & 13.87 & 36.91\\
            ASER\pub{AAAI2021}\cite{shim2021online} & 21.42 & 54.30 & 16.42 & 60.55 & 6.02 & 55.01 & 15.23 & 61.06 & 8.74 & 64.79 & 9.74 & 64.87\\
            \hline
            ER-ACE\pub{ICLR2021}\cite{caccia2021new} & 48.56 & 13.82 & 50.89 & 13.13 & 42.47 & 17.32 & 45.01 & 14.68 & 38.92 & 24.72 & 38.51 & 16.71\\   
			$+\textbf{\texttt{Ours}}$ &\reshl{\textbf{49.35}}{0.79} & \reshlup{\textbf{17.90}}{4.08} & \reshl{\textbf{54.22}}{3.33} & \reshldown{\textbf{12.01}}{1.12} & \reshl{\textbf{43.57}}{1.10} & \reshlup{\textbf{18.04}}{0.72} & \reshl{\textbf{45.94}}{0.93} & \reshlup{\textbf{15.15}}{0.47} & \reshl{\textbf{40.12}}{1.20} & \reshldown{\textbf{24.03}}{0.69} & \reshl{\textbf{40.16}}{1.65} & \reshldown{\textbf{16.22}}{0.49}\\

            \hline
            CLS-ER\pub{ICLR2022}\cite{arani2022learning} & 47.92 & 36.54 & 48.29 & 23.01 & 45.11 & 34.67 & 45.84 & 35.24 & 39.04 & 41.48 & 35.97 & 41.12\\
			$+\textbf{\texttt{Ours}}$ &\reshl{\textbf{48.91}}{0.99} & \reshldown{\textbf{35.15}}{1.39} & \reshl{\textbf{56.99}}{8.70} & \reshldown{\textbf{17.53}}{5.48} & \reshldownred{\textbf{43.91}}{1.20} & \reshlup{\textbf{37.45}}{2.78} & \reshldownred{\textbf{45.31}}{0.53} & \reshlup{\textbf{35.41}}{0.17} & \reshl{\textbf{40.68}}{1.64} & \reshldown{\textbf{38.21}}{3.27} & \reshl{\textbf{37.18}}{1.21} & \reshldown{\textbf{39.58}}{1.54}\\
   
            \hline
            PCR\pub{CVPR2023}\cite{lin2023pcr} & 47.63 & 13.21 & 65.55 & 14.47 & 46.26 & 15.72 & 46.19 & 16.02 & 38.95 & 23.76 & 37.76 & 18.85\\
			$+\textbf{\texttt{Ours}}$ &\reshl{\textbf{49.96}}{2.33} & \reshldown{\textbf{12.83}}{0.38} & \reshl{\textbf{68.20}}{2.65} & \reshldown{\textbf{7.73}}{6.74} & \reshl{\textbf{48.75}}{2.49} & \reshldown{\textbf{13.17}}{2.55} & \reshl{\textbf{47.91}}{1.72} & \reshldown{\textbf{15.62}}{0.40} & \reshl{\textbf{40.69}}{1.74} & \reshldown{\textbf{20.63}}{3.13} & \reshl{\textbf{38.64}}{0.88} & \reshldown{\textbf{14.65}}{4.20}\\
			\hline
	\end{tabular}}
	\label{main_result}
\vspace{-2mm}
\end{table*}

\begin{table*}[t]
	\centering
	\caption{{Quantitative results using different buffer sizes on CIFAR100~\cite{krizhevsky2009learning} and Mini-ImageNet~\cite{vinyals2016matching}}}
	\setlength\tabcolsep{6pt}
	\renewcommand\arraystretch{1.1}
	\resizebox{\textwidth}{!}{
		\begin{tabular}{c|cc|cc|cc|cc|cc|cc}
			\thickhline
			Datasets & \multicolumn{6}{c|}{CIFAR100} & \multicolumn{6}{c}{Mini-ImageNet} \\
            \hline
            Buffer Size & \multicolumn{2}{c|}{1000} & \multicolumn{2}{c|}{5000} & \multicolumn{2}{c|}{10000} & \multicolumn{2}{c|}{1000} & \multicolumn{2}{c|}{5000} & \multicolumn{2}{c}{10000} \\
            \hline
            Methods & \textit{Acc. ($\uparrow$ )} & \textit{Fgt. ($\downarrow$ )} & \textit{Acc. ($\uparrow$ )} & \textit{Fgt. ($\downarrow$ )} & \textit{Acc. ($\uparrow$ )} & \textit{Fgt. ($\downarrow$ )} & \textit{Acc. ($\uparrow$ )} & \textit{Fgt. ($\downarrow$ )} & \textit{Acc. ($\uparrow$ )} & \textit{Fgt. ($\downarrow$ )} & \textit{Acc. ($\uparrow$ )} & \textit{Fgt. ($\downarrow$ )} \\
            \hline
            ER-ACE\pub{ICLR2021} & 30.12 & 37.11 & 42.32 & 19.50 & 48.56 & 13.82 & 36.41 & 27.02 & 40.82 & 22.29 & 45.01 & 14.68 \\   
            $+\textbf{\texttt{Ours}}$ & \reshl{30.54}{0.42} & \reshldown{34.50}{2.61} & \reshl{43.40}{1.08} & \reshlup{24.17}{4.67} & \reshl{\textbf{49.35}}{0.79} & \reshlup{\textbf{17.90}}{4.08} & \reshl{38.21}{1.80} & \reshldown{26.54}{0.48} & \reshl{42.15}{1.33} & \reshldown{20.63}{1.66} & \reshl{\textbf{45.94}}{0.93} & \reshlup{\textbf{15.15}}{0.47} \\
            CLS-ER\pub{ICLR2022} & 22.44 & 66.29 & 41.24 & 42.20 & 47.92 & 36.54 & 18.89 & 68.01 & 39.54 & 44.87 & 45.84 & 35.24 \\
            $+\textbf{\texttt{Ours}}$ & \reshl{22.97}{0.53} & \reshldown{66.01}{0.28} & \reshl{42.26}{1.04} & \reshldown{41.41}{0.79} & \reshl{\textbf{48.91}}{0.99} & \reshldown{\textbf{35.15}}{1.39} & \reshldownred{18.56}{0.33} & \reshlup{69.96}{1.95} & \reshl{40.35}{0.81} & \reshldown{44.11}{0.76} & \reshldownred{45.31}{0.53} & \reshlup{35.41}{0.17} \\
            PCR\pub{CVPR2023}\cite{lin2023pcr} & 29.24 & 24.66 & 43.31 & 16.18 & 47.63 & 13.21 & 24.37 & 42.63 & 40.86 & 21.66 & 46.19 & 16.02 \\
            $+\textbf{\texttt{Ours}}$ & \reshl{30.80}{1.56} & \reshldown{22.38}{2.28} & \reshl{45.14}{1.83} & \reshldown{16.12}{0.06} & \reshl{49.96}{2.33} & \reshldown{12.83}{0.38} & \reshl{26.94}{2.57} & \reshldown{38.67}{3.96} & \reshl{42.54}{1.68} & \reshldown{20.01}{1.65} & \reshl{47.91}{1.72} & \reshldown{15.62}{0.40} \\
			\hline
	\end{tabular}}
	\label{main_result:2}
\vspace{-2mm}
\end{table*}

\begin{figure*}[h!]
  \centering
      \includegraphics[width=1 \linewidth]{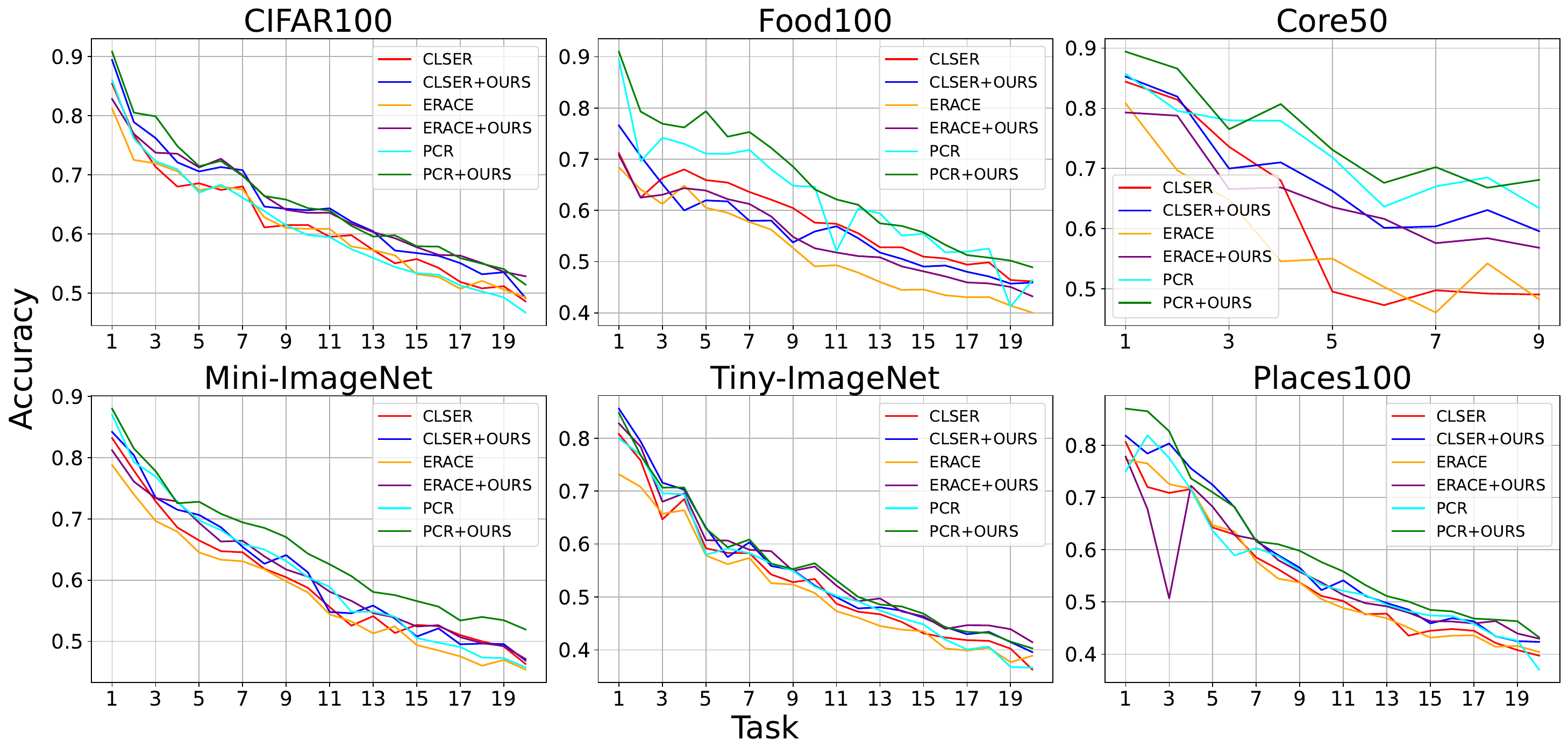}
\captionsetup{aboveskip=-1mm} 
\caption{The average class accuracy after +k task batches since the moment a task appeared.}
    \label{fig:task-batch}
\vspace{-3mm}
\end{figure*}

\noindent \textbf{Evaluation Metrics.} We employ three key metrics: \textit{Average End Accuracy (Acc.)}, \textit{Average Forgetting (Fgt.)}, and \textit{Average Retention Rate}. For details regarding the calculations, please refer to the supplementary material.

\noindent \textbf{Reproducibility.} Our algorithm has been implemented using the PyTorch. All experiments were performed on a setup comprising a NVIDIA Tesla A100 GPU with 40GB of memory. To ensure the reproducibility of our results, we plan to make our code publicly available.


\subsection{Results} \label{label:re}

\newlength{\hfigsep}
\setlength{\hfigsep}{0.3mm} 
\newlength{\vfigsep}
\setlength{\vfigsep}{0.3mm} 
\textbf{Qualitative Results on Benchmark Datasets.} In Table~\ref{main_result}, we present quantitative results evaluating the performance of various continual learning methods across six benchmark datasets: CIFAR100, Core50, Food100, Mini-ImageNet, Places100, and Tiny-ImageNet. The evaluation metrics used are average end accuracy (Acc.) and forgetting (Fgt.). 

The methods evaluated PCR~\cite{lin2023pcr} and version enhanced with our balanced retrieval (denoted as "+ Ours"), with ER~\cite{chaudhry2019tiny}, A-GEM~\cite{chaudhry2018efficient}, DER++~\cite{buzzega2020dark}, ASER~\cite{shim2021online}, ER-ACE~\cite{caccia2021new}, and CLS-ER~\cite{arani2022learning}. Additionally, we wanted to evaluate how our balanced retrieval will behave when combined with memory-based methods not relying on proxies, so we have augmented ER-ACE and CLS-ER with it.

The results demonstrate that our proposed method yields substantial improvements when applied to PCR~\cite{lin2023pcr}, achieving the highest gains in both accuracy and reduced forgetting. This highlights the effectiveness of our approach in mitigating catastrophic forgetting while maintaining high accuracy. Additionally, our method performs well across most of the datasets compared to the baseline methods. Interestingly, our retrieval boosts performance of both ER-ACE and CLS-ER, although not as significantly as in the case of PCR. This shows that exploring the interplay between gradient-aligned and gradient-conflicting samples during retrieval may be also a promising direction for methods based on cross-entropy loss. 

For instance, on CIFAR100, PCR~\cite{lin2023pcr}+Ours achieves an accuracy of 49.96\% and forgetting of 12.83\%, compared to the baseline PCR~\cite{lin2023pcr}'s accuracy of 47.63\% and forgetting of 13.21\%. Similar trends are observed across all other datasets, validating the robustness and generalization capabilities of our method. 

In Table \ref{main_result:2},we present an evaluation of the enhanced versions of ER-ACE~\cite{caccia2021new}, CLS-ER~\cite{arani2022learning}, and PCR~\cite{lin2023pcr}, each integrated with our proposed method (referred to as "+ Ours"), on CIFAR-100 and Mini-ImageNet across three different buffer sizes. Overall, the use of balanced sampling demonstrates a consistent and stable improvement in performance.

\noindent \textbf{Analysis of the Task-Batch Performance}
Figure~\ref{fig:task-batch} shows the average accuracy after each task is introduced incrementally. This visual representation allows us to analyze the stability of the evaluated methods and their response to the increasing model size as more tasks are stored and remembered. Our proposed method demonstrates significantly improved stability compared to the baseline approaches, maintaining superior predictive accuracy regardless of the number of classes. In the plots for Core50, Food100, and Mini-ImageNet, it is evident that PCR~\cite{lin2023pcr}+OURS consistently outperforms the others. A detailed comparison reveals that all three promoted models show some enhancement over their baselines version.

\begin{figure*}[t]
  \centering
      \includegraphics[width=1 \linewidth]{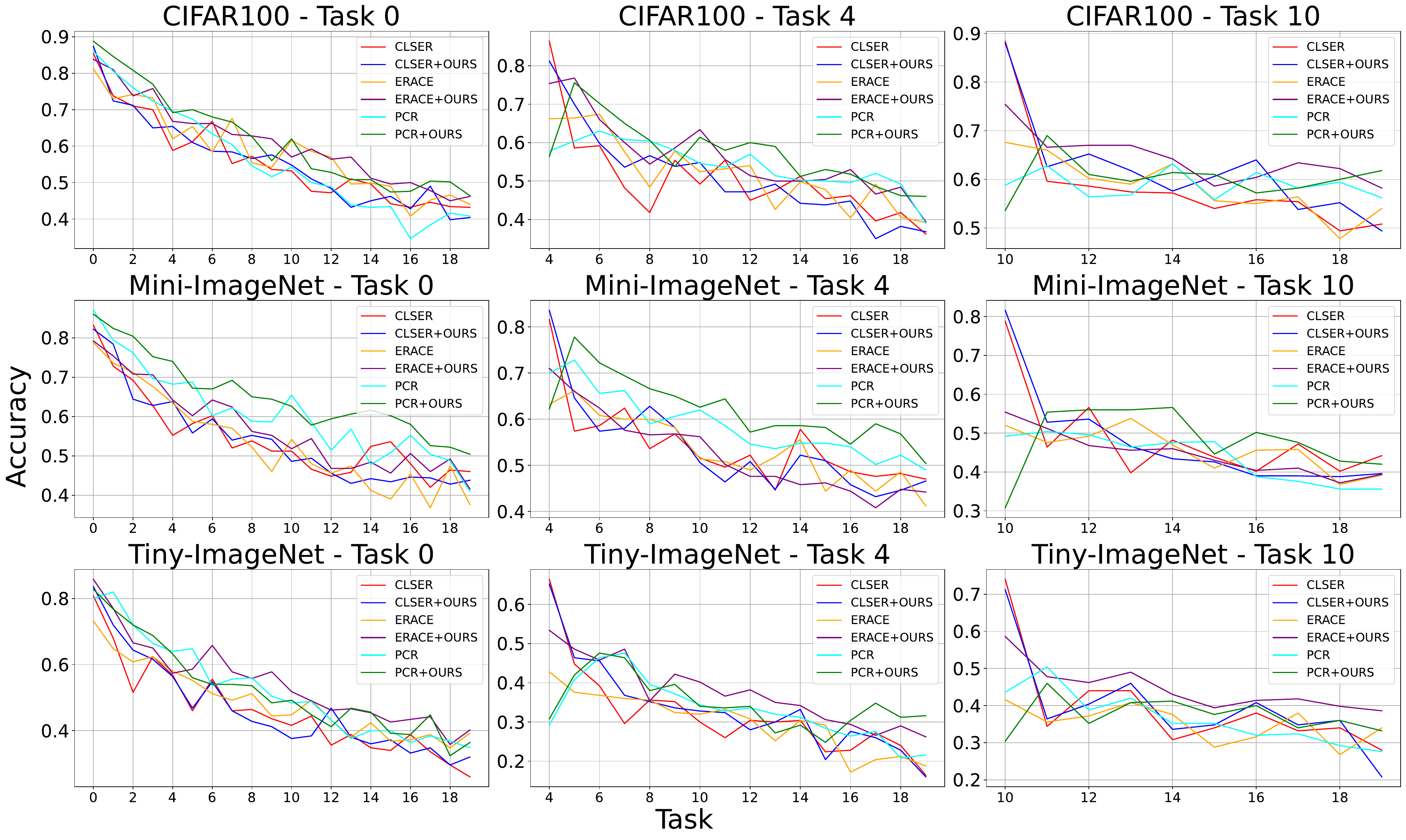}
\captionsetup{aboveskip=-0.5mm} 
\caption{The average accuracy for selected classes for different models after subsequent task batches.}
    \label{fig:class}
\vspace{-2mm}
\end{figure*}

\begin{figure}[t]
  \centering
      \includegraphics[width=1 \linewidth]{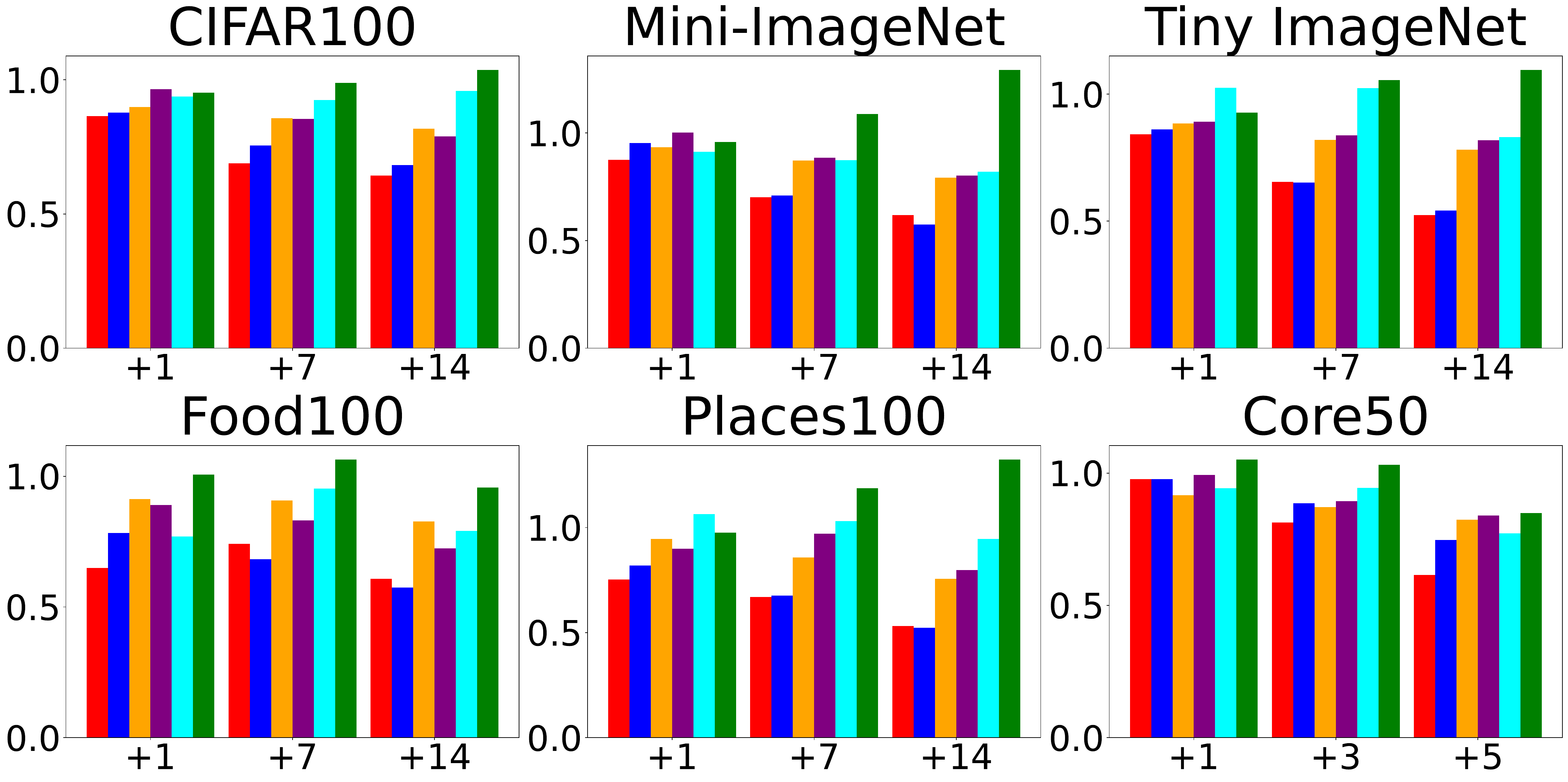}
\caption{The average retention rate after +k task batches since the moment a task appeared
for: {\color{red}\rule{1.4ex}{1.4ex}} CLSER, {\color{blue}\rule{1.4ex}{1.4ex}} CLSER+OURS, {\color{orange}\rule{1.4ex}{1.4ex}} ER-ACE, {\color{purple}\rule{1.4ex}{1.4ex}} ER-ACE+OURS, {\color{cyan}\rule{1.4ex}{1.4ex}} PCR, {\color{green}\rule{1.4ex}{1.4ex}} PCR+OURS.}
\vspace{-7mm}
    \label{fig:retention}
\end{figure}

\begin{table*}[t]
\centering
\caption{Effect of different sampling ratios (bounded between pure MIR and pure IMIR) on final accuracy.}
\begin{adjustbox}{width=0.95\textwidth}  
\begin{tabular}{c|cccccccccccc}
\toprule
\textbf{MIR/IMIR} & \textbf{10:0} & \textbf{9:1} & \textbf{8:2} & \textbf{7:3} & \textbf{6:4} & \textbf{5:5} & \textbf{4:6} & \textbf{3:7} & \textbf{2:8} & \textbf{1:9} & \textbf{0:10} & \textbf{random} \\
\midrule
\textbf{CIFAR100} & 49.08 & 48.34 & 49.29 & 49.87 & 50.06 & 49.96 & 49.91 & 49.92 & 49.48 & 48.77 & 49.09 & 47.63 \\
\midrule
\textbf{Mini Imagenet} & 47.04 & 46.24 & 47.23 & 47.79 & 47.81 & 47.91 & 47.57 & 47.80 & 47.55 & 46.61 & 47.08 & 46.19 \\
\bottomrule
\end{tabular}
\end{adjustbox}
\label{table:ablation}
\end{table*}

\begin{figure*}[h!]
  \centering
      \includegraphics[width=1 \linewidth]{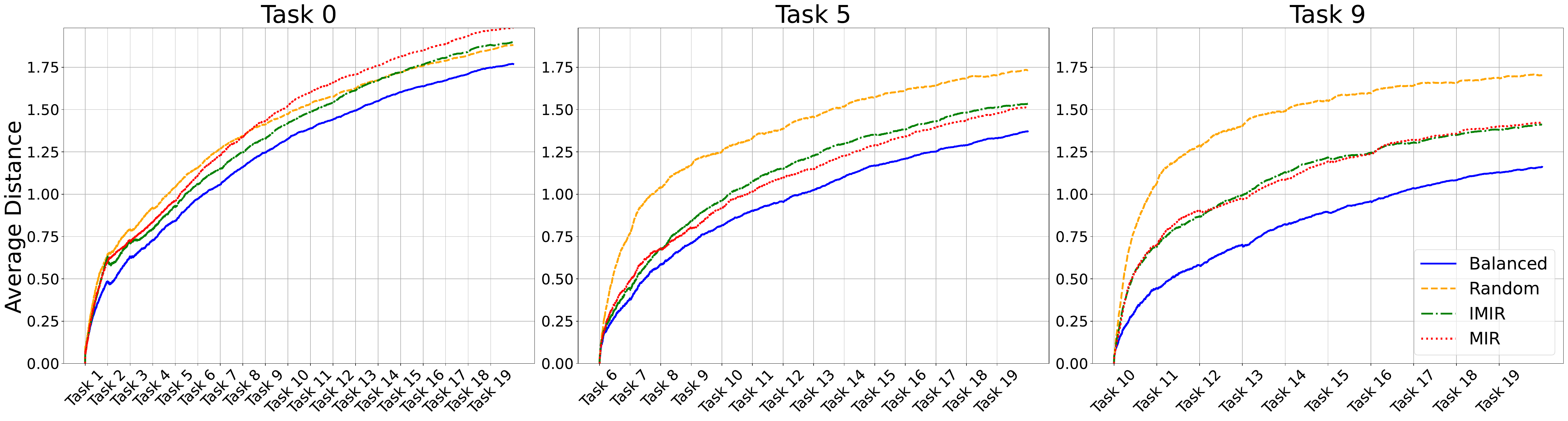}
\setlength{\abovecaptionskip}{-4mm} 
\caption{The average proxy drift for selected tasks from CIFAR100.}
    \label{fig:proxys-drift}
\vspace{-2mm}
\end{figure*}
\begin{figure*}[h!]
  \centering
      \includegraphics[width=1 \linewidth]{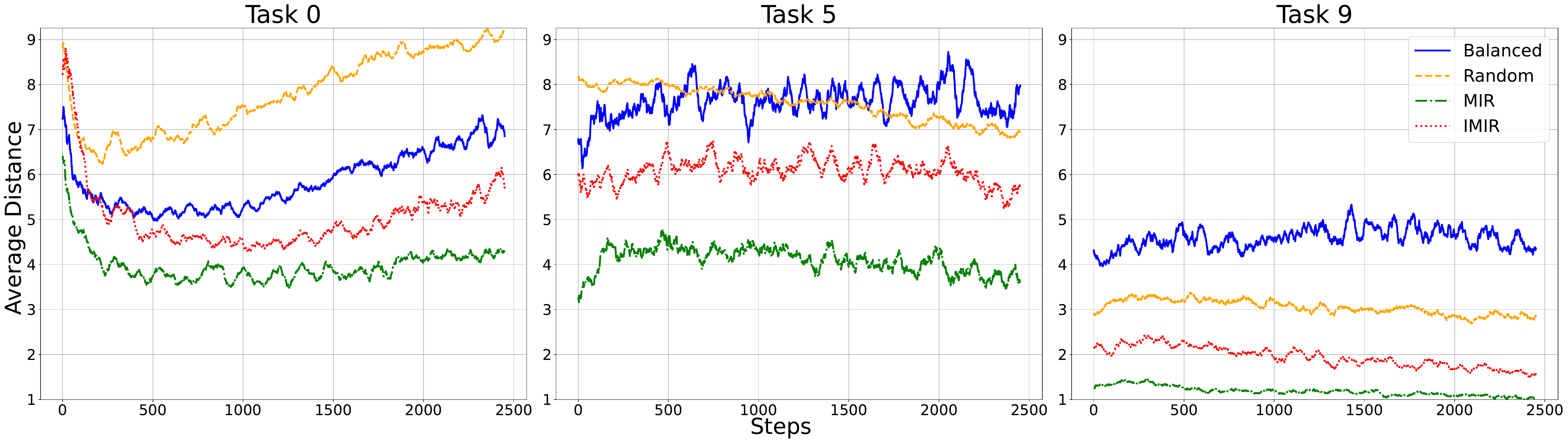}
\setlength{\abovecaptionskip}{-4mm} 
\caption{The average inner distance of retrieved samples during the training of selected tasks from CIFAR100.}
    \label{fig:inner-distance}
\vspace{-2mm}
\end{figure*}
\begin{figure*}[h!]
  \centering
      \includegraphics[width=1 \linewidth]{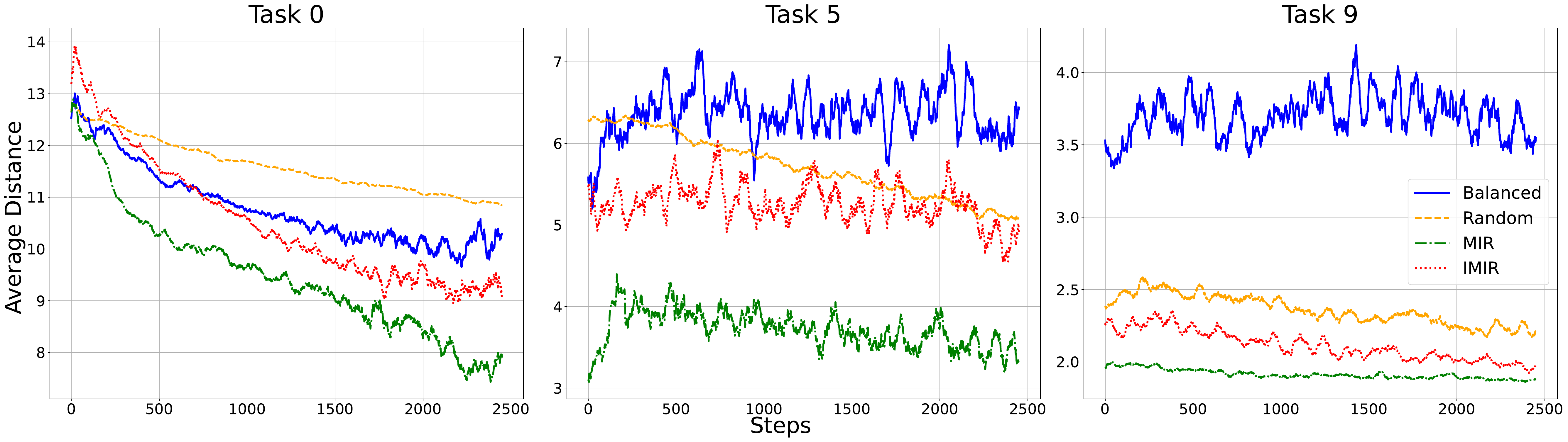} 
\setlength{\abovecaptionskip}{-4mm} 
\caption{The average distance from retrieved samples to their corresponding proxies on selected tasks from CIFAR100.}
\vspace{-5mm}
    \label{fig:samples_to_proxys}
\end{figure*}

\noindent \textbf{Analysis of the Class-Based Performance} Figure~\ref{fig:class} depicts the accuracy per batch on selected classes. This allows us to understand how the appearance of new classes affects the performance on previously seen ones. We plot the comparison of CLSER, ER-ACE, and PCR with and without our balanced sampling. In most cases (especially for early tasks), we can see that the addition of our balanced sampling leads to slower decrease of accuracy over time and thus to greater robustness to catastrophic forgetting. Interestingly, for some cases (e.g., Places100 or Mini-ImageNet) we can see that the accuracy stabilizes over time, leading to a semi-plateau, where approaches without our sampling keep decreasing their accuracy. This shows that our balanced sampling may have an underlying long-term impact on stabilization of task performance over time and could be beneficial for lifelong data streams. This behavior is observed mostly for oldest tasks, that had been in the memory buffer for long periods of time.



\noindent \textbf{Analysis of the Retention of Information.} Figure \ref{fig:retention} illustrates the average retention of task-specific knowledge upon the introduction of +k new tasks. This figure is crucial for assessing the ability of each model to maintain its knowledge base while integrating new information. Ideally, a model should consistently perform well on all previously encountered tasks, regardless of the subsequent addition of new tasks. Our findings indicate that PCR~\cite{lin2023pcr}+OURS significantly outperforms the other two state-of-the-art methods. Moreover, the integration of our proposed module leads to a notable enhancement in average accuracy as the number of new tasks increases.

The enhanced PCR~\cite{lin2023pcr} method demonstrates an interesting behavior where its retention rate can sometimes exceed 1. This happens because, during the training of new tasks, some tasks initially show lower accuracy due to the limitations of the dual retrieval mechanism. However, their accuracy improves during subsequent training phases. This suggests that balanced sampling may lead to some underfitting at the early stages of training for certain tasks. For example, in the case of CIFAR-100, the initial accuracies of PCR~\cite{lin2023pcr}+Ours for each task are 0.888, 0.724, 0.772, 0.658, 0.564, 0.662, 0.584, 0.452, 0.538, 0.548, 0.536, 0.340, 0.394, 0.496, 0.376, 0.580, 0.352, 0.300, and 0.428. In comparison, the final accuracies are 0.464, 0.496, 0.548, 0.622, 0.460, 0.664, 0.560, 0.460, 0.536, 0.592, 0.618, 0.376, 0.482, 0.442, 0.496, 0.552, 0.436, 0.438, and 0.550, respectively. The overall average retention ratio calculated is 1.019.


\noindent \textbf{Robustness Analysis to Proxy Drift.} In Fig.~\ref{fig:proxys-drift}, we present the progression of proxy drift over time across three selected tasks from CIFAR-100, with each task's drift averaged over its respective classes. Notably, random retrieval experiences the highest degree of drift, underscoring its vulnerability. In contrast, both MIR and IMIR demonstrate enhanced robustness, particularly as tasks progress. Our proposed balanced sampling strategy significantly mitigates proxy drift across all tasks, an effect that becomes even more pronounced when analyzing drift at the class level rather than the task level (see Appendix for detailed class-wise results). This improvement stems from retrieving a balanced mix of aligned and conflicting gradients from the buffer, which facilitates a more comprehensive data representation of prior tasks and reduces drift magnitude.

\noindent \textbf{Analysis of the diversity of retrieved instances.} To further validate that our proposed balanced sampling achieves an improved trade-off between adapting to new data and retaining prior knowledge, we assess the diversity of retrieved samples. Fig.~\ref{fig:inner-distance} depicts the average distance among embeddings retrieved from the buffer, while Fig.~\ref{fig:samples_to_proxys} shows the average distance between retrieved instances and their corresponding class proxies. The proposed sampling strategy selects notably more diverse instances, particularly in later tasks. Drawing on findings from \cite{Wang:2020al}, we infer that the diversity among buffer-retrieved instances and their expanded coverage of decision space relative to class proxies enhance both adaptation and retention, reducing the bias observed in MIR/IMIR, which tend to prioritize either new task adaptation or knowledge retention exclusively.

\subsection{Effects of sampling ratios}\label{ablation}
\vspace{-2mm}
In this part, we investigate the impact of varying sampling ratios between these two strategies on model performance. Our experimental design incorporates adaptive adjustments to the proportions of samples selected in each training batch, as detailed in Table~\ref{table:ablation}. The results demonstrate that a balanced sampling approach significantly enhances model performance, with optimal gains observed when the sampling distribution was approximately an even split (around 5:5). Notably, performance under extreme ratio conditions (such as 10:0 and 0:10), where samples were drawn exclusively from one strategy, served as important baselines. These cases clearly showed that relying solely on a single strategy leads to sub-optimal performance due to sampling bias and insufficient coverage of representative instances for each class or task. This highlights the importance of a balanced sampling approach, which allows for the extraction of more effective samples from the buffer for training.

\section{Conclusions and Future Works}\label{conclusion}

\noindent \textbf{Summary.} In this paper, we addressed the challenge of catastrophic forgetting in contrastive continual learning by proposing a novel balanced instance retrieval strategy from the buffer that leverages both gradient-aligned and gradient-conflicting samples. Our approach was designed to enhance stability in shared representations while preserving past knowledge, effectively mitigating proxy drift. Through theoretical analysis, we showed that gradient-aligned samples help stabilize embeddings shared across tasks, while gradient-conflicting samples play a critical role in knowledge retention. We empirically showed that our approach improves the diversity of retrieved instances, reduces proxy drift, and surpasses existing methods in both retention and adaptation to new tasks. 

\noindent \textbf{Future works.} We plan to further analyze what factors have the highest impact on drift occurrence in contrastive learning, as well as focus on enhancing our sampling strategy with more explicit drift prevention mechanisms. Furthermore, we plan to extend our framework to continual learning with extremely limited access to ground truth.
{
\small
\bibliographystyle{ieeenat_fullname}

}

\clearpage
\setcounter{page}{1}
\addtocounter{section}{5}
\maketitlesupplementary


\section{Gradient alignment visualization}
The motivation for selecting samples from both ends of the distribution of the criterion $\mathcal{C}$ is that the gradients of samples with the highest $\mathcal{C}$ (selected by MIR) are misaligned with the gradients of the training batch. Therefore, to reinforce alignment, we also select samples with the lowest $\mathcal{C}$. However, this reasoning stems from the approximation of $\mathcal{C}$. In Fig.~7, we present the calculated gradient alignment of the loss function for a batch with samples retrieved by MIR and IMIR across tasks 1, 3, 5, and 7, with other tasks following the same trend. The samples selected by MIR exhibit significantly less aligned gradients compared to those selected by IMIR, which supports the proposed motivation.

\begin{figure}[h!]
  \centering
      \includegraphics[width=1 \linewidth]{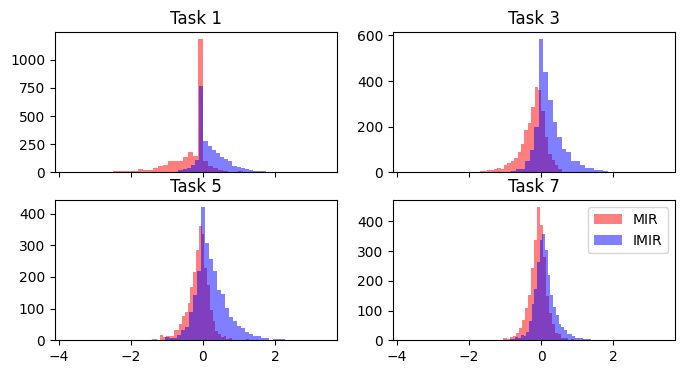}
\setlength{\abovecaptionskip}{-4mm} 
\caption{Gradient alignment of samples selected by MIR (red) and negative MIR (blue) with the batch gradients.}
    \label{fig:histograms}
\vspace{-2mm}
\end{figure}

\section{Metrics}
\textit{Average End Accuracy} is the mean accuracy across all tasks after the final training session:
\vspace{-2mm}
\begin{equation}
Acc. = \frac{1}{T} \sum_{t=1}^T A_{T,t},
\end{equation}
where \( T \) is the total number of tasks, and \( A_{T,t} \) is the accuracy on task \( t \) after training on the final task.

\textit{Average Forgetting} measures the average reduction in accuracy for previously learned tasks:
\vspace{-2mm}
\begin{equation}
Fgt. = \frac{1}{T-1} \sum_{t=1}^{T-1} f_t, \quad f_t = \max_{i \leq t} \{ A_{i,t} \} - A_{T,t},
\end{equation}
\vspace{-4mm}

where \( f_t \) is the forgetting on task \( t \), \( A_{i,t} \) is the accuracy on task \( t \) after training on task \( i \), and \( \max_{i \leq t} \{ A_{i,t} \} \) is the maximum accuracy achieved on task \( t \) before training on the final task.

\textit{Average Retention Rate} captures the model's ability to retain knowledge, and is given by:
\vspace{-2mm}
\begin{equation}
ARR = \frac{1}{T-1} \sum_{t=1}^{T-1} r_t, \quad r_t = \frac{A_{T,t}}{\max_{i \leq t} A_{i,t}},
\end{equation}
where \( r_t \) is the retention rate for task \( t \), \( A_{T,t} \) is the accuracy on task \( t \) after training on the final task \( T \), and \( \max_{i \leq t} A_{i,t} \) is the maximum accuracy achieved on task \( t \) before training on the final task.

\section{Datasets}

\noindent \textbullet \ \textit{CIFAR100 \cite{krizhevsky2009learning}} consists of 60,000 color images of 32x32 pixels, divided into 100 classes with 600 images each. The dataset is split into 50,000 training images and 10,000 testing images.

\noindent \textbullet \ \textit{Core50 \cite{lomonaco2017core50}} is tailored for continual learning and object recognition, comprising 50,000 images of 50 objects divided into 10 categories. The dataset features variability in background, lighting, and viewpoint, with images captured at a resolution of 350x350 pixels. For experimental purposes, it is segmented into 40,000 training images and 10,000 testing images.

\noindent \textbullet \ \textit{Food100} is a subset of Food-101~\cite{bossard2014food}, which includes 100 food classes with 50,000 training and 10,000 testing samples, independent from ImageNet~\cite{deng2009imagenet}. Each image has a maximum side length of 512 pixels, catering to detailed, domain-specific food recognition tasks.

\noindent \textbullet \ \textit{Mini-ImageNet \cite{vinyals2016matching}} is a condensed version of the ImageNet \cite{deng2009imagenet}. It includes 60,000 training and 20,000 testing images, distributed across 100 classes, with each image resized to 84x84 pixels.

\noindent \textbullet \ \textit{Places100} is a curated subset of the Places365~\cite{zhou2017places}, featuring 100 diverse scene categories. It comprises 45,000 training and 10,000 testing images, providing a focused yet comprehensive environment for training scene recognition models. Each image is typically 256x256 pixels.

\noindent \textbullet \ \textit{Tiny-ImageNet \cite{le2015tiny}} scales down the ImageNet \cite{deng2009imagenet} for more accessible image classification challenges. We select a subset with 100 classes from it. It features 50,000 training and 5000 testing images, with each image presented in 64x64 pixels. 
Having the supplementary compiled together with the main paper means that:
%

\section{Hyperparameters}

Table~\ref{tab:hyperparameters} shows the details of hyperparameters used in our experiments.

\begin{table}[h!]
\centering
\caption{Hyperparameter Settings}
\label{tab:hyperparameters}
\renewcommand{\arraystretch}{1.1} 
\begin{tabular}{>{\raggedright\arraybackslash\scriptsize}m{4cm} | >{\raggedright\arraybackslash\scriptsize}m{3.5cm}}
\toprule
\textbf{Hyperparameter} & \textbf{Value} \\
\midrule
\rowcolor{gray!20} Optimizer & SGD \\
Learning Rate & 0.1 \\
\rowcolor{gray!20} Training Batch Size & 10 \\
Testing Batch Size & 128 \\
\rowcolor{gray!20} Number of Tasks & 20 \\
Instances Retrieved Per Batch & 10 \\
\rowcolor{gray!20} Weight Initialization & Xavier Uniform \\
CLSER Consistency Regularization Weight & 0.1 \\
\rowcolor{gray!20} CLSER Stable Model Update Freq & 0.7 \\
CLSER Stable Model Alpha & 0.999 \\
\rowcolor{gray!20} CLSER Plastic Model Update Freq & 0.9 \\
CLSER Plastic Model Alpha & 0.999 \\
\rowcolor{gray!20} DER++ Alpha & 0.5 \\
DER++ Beta & 0.5 \\
\bottomrule
\end{tabular}
\end{table}

\section{Comparison with other instance selection methods.}
The main text of our paper includes a comparison with MIR (SOTA instance retrieval), as it is the best performing reference method for proxy-based continual learning. Here, we present an additional comparison with other recent instance retrieval methods: ASER\cite{shim2021online}/GSS\cite{AljundiLGB19}/Memory matching\cite{MAI202228}/MIR\cite{aljundi2019online}/Random Retrieval. Table \ref{tab:comparison_results} depicts how PCR performs when combined with ASER/GSS/Memory matching/MIR/Random versus our solution. We select results for CIFAR100 and Mini-ImageNet, as they are representative of the performance trends over all 6 datasets. 


%
\vspace{-2mm}
\begin{table}[H]
\centering
\caption{Comparison of methods on CIFAR100 and Mini-ImageNet datasets.}
\label{tab:comparison_results}
\renewcommand{\arraystretch}{1.2}
\scriptsize  
\resizebox{0.45\textwidth}{!}{  
\begin{tabular}{l | c | c}
\toprule
\textbf{Method} & \textbf{CIFAR100} & \textbf{Mini-ImageNet} \\
\midrule
\rowcolor{gray!20} PCR+OURS & \textbf{49.96} & \textbf{47.91} \\
PCR+MIR\cite{aljundi2019online} & 49.08 & 47.04 \\
\rowcolor{gray!20} PCR+GSS\cite{AljundiLGB19} & 5.61 & 5.32 \\
PCR+ASER\cite{shim2021online} & 21.92 & 27.80 \\
\rowcolor{gray!20} PCR+Memmatch\cite{MAI202228} & 42.04 & 45.93 \\
PCR+Random & 47.63 & 46.19 \\
\bottomrule
\end{tabular}
}
\end{table}

\section{Statistical tests}
A Wilcoxon signed-rank test \cite{Demsar:2006} has shown in Table \ref{tab:statistical_tests} to measure the statistical significance of the results. This pairwise statistical test compares two classifiers over multiple datasets (6 datasets in our case). All results were obtained using 5-10 repetitions, and p-values were measured for both Accuracy and Forgetting measures, with a significance level of 0.05 (see Table \ref{tab:statistical_tests}). As we can see, the proposed balanced retrieval leads to statistically significant improvements for ER-ACE, CLS-ER, and PCR, with PCR showing the strongest statistical difference. This further supports our claims that our proposed instance retrieval method works particularly well with proxy-based continual learning. 

\begin{table}[h!]
\centering
\caption{Results of Wilcoxon signed-rank test for comparison over multiple datasets with significance level = 0.05}
\label{tab:statistical_tests}
\renewcommand{\arraystretch}{1.1} 
\begin{tabular}{>{\raggedright\arraybackslash\scriptsize}m{3cm} | >{\centering\arraybackslash\scriptsize}m{2cm} | >{\centering\arraybackslash\scriptsize}m{2cm}}
\toprule
\textbf{Comparison} & \textbf{Acc. ($p$-values)} & \textbf{Fgt. ($p$-values)} \\
\midrule
\rowcolor{gray!20} ER-ACE+Ours vs. ER-ACE & 0.0000 & 0.0177 \\
CLS-ER+Ours vs. CLS-ER & 0.0068 & 0.0082 \\
\rowcolor{gray!20} PCR+Ours vs. PCR & 0.0000 & 0.0000 \\
\bottomrule
\end{tabular}
\end{table}

We employed McNemar's test \cite{Demsar:2006} to evaluate whether the differences between the proposed balanced retrieval and reference retrieval methods are statistically significant on individual datasets. Table~\ref{tab:statistical_tests2} summarizes the results, obtained through 5–10 repetitions, with p-values computed separately for CIFAR100 and Mini-ImageNet at a significance level of 0.05. The results indicate that the proposed balanced retrieval, when paired with the PCR algorithm, achieves statistically significant performance improvements over all reference methods. This further reinforces our claim that proxy-based continual learning benefits from a specialized retrieval strategy that enhances adaptation to new classes while improving knowledge retention by implicitly mitigating proxy drift.

\begin{table}[h!]
\centering
\caption{Results of McNemar's test for comparison between retrieval methods over individual datasets with significance level = 0.05}
\label{tab:statistical_tests2}
\renewcommand{\arraystretch}{1.1} 
\begin{tabular}{>{\raggedright\arraybackslash\scriptsize}m{3cm} | >{\centering\arraybackslash\scriptsize}m{2cm} | >{\centering\arraybackslash\scriptsize}m{2cm}}
\toprule
\textbf{Comparison} & \textbf{CIFAR100 ($p$-values)} & \textbf{Mini-ImageNet ($p$-values)} \\
\midrule
\rowcolor{gray!20} PCR+OURS vs. PCR+MIR & 0.0309 & 0.0311 \\
PCR+OURS vs. PCR+GSS & 0.0000 & 0.0000 \\
\rowcolor{gray!20} PCR+OURS vs. PCR+ASER & 0.0000 & 0.0000 \\
PCR+OURS vs. PCR+Memmatch & 0.0012 & 0.0126 \\
\rowcolor{gray!20} PCR+OURS vs. PCR+Random & 0.0197 & 0.0208 \\
\bottomrule
\end{tabular}
\end{table}

\begin{table*}[!t]
\centering
\caption{Effect of different sampling ratios (bounded between pure MIR and pure IMIR) on \textbf{Average End Forgetting}.}
\begin{adjustbox}{width=0.95\textwidth}  
\begin{tabular}{c|cccccccccccc}
\toprule
\textbf{MIR/IMIR} & \textbf{10:0} & \textbf{9:1} & \textbf{8:2} & \textbf{7:3} & \textbf{6:4} & \textbf{5:5} & \textbf{4:6} & \textbf{3:7} & \textbf{2:8} & \textbf{1:9} & \textbf{0:10} & \textbf{random} \\
\midrule
\textbf{CIFAR100} & 12.67 & 13.09 & 12.79 & 12.43 & 13.06 & 12.83 & 13.19 & 12.19 & 12.47 & 12.83 & 12.15 & 13.21 \\
\midrule
\textbf{Mini-ImageNet} & 14.86 & 16.05 & 15.48 & 15.14 & 15.44 & 15.62 & 15.81 & 14.73 & 14.91 & 15.55 & 13.11 & 16.02 \\
\bottomrule
\end{tabular}
\end{adjustbox}
\label{table:ablation2}
\end{table*}

\section{Forgetting for MIR and IMIR}

In the main paper, Table 3 analyzed the effect of varying sampling ratios (ranging from pure MIR to pure IMIR) on final accuracy, demonstrating that the proposed balanced retrieval outperforms either method individually. To further support this observation, Table~\ref{table:ablation2} examines the influence of sampling ratios on Average End Forgetting. While accuracy metrics show similar performance for both MIR and IMIR, this suggests that their combined strength arises from differing impacts on proxy-based continual learning. Table~\ref{table:ablation2} provides additional insights, revealing that pure IMIR significantly reduces forgetting compared to pure MIR. This supports the conclusion that the superior performance of the balanced retrieval approach stems from its ability to balance between instances, optimizing both adaptation to new classes and retention of previously learned tasks.

\section{Additional plots for proxy drift, retrieval diversity, and accuracy}

Due to space constraints in the main paper, we presented a limited selection of plots illustrating proxy drift and retrieval diversity. Here, we provide additional visualizations to further substantiate our claim that the proposed balanced retrieval strategy effectively mitigates proxy drift while enhancing the diversity of retrieved instances from the buffer.

\begin{figure*}[h!]
  \centering
      \includegraphics[width=1 \linewidth]{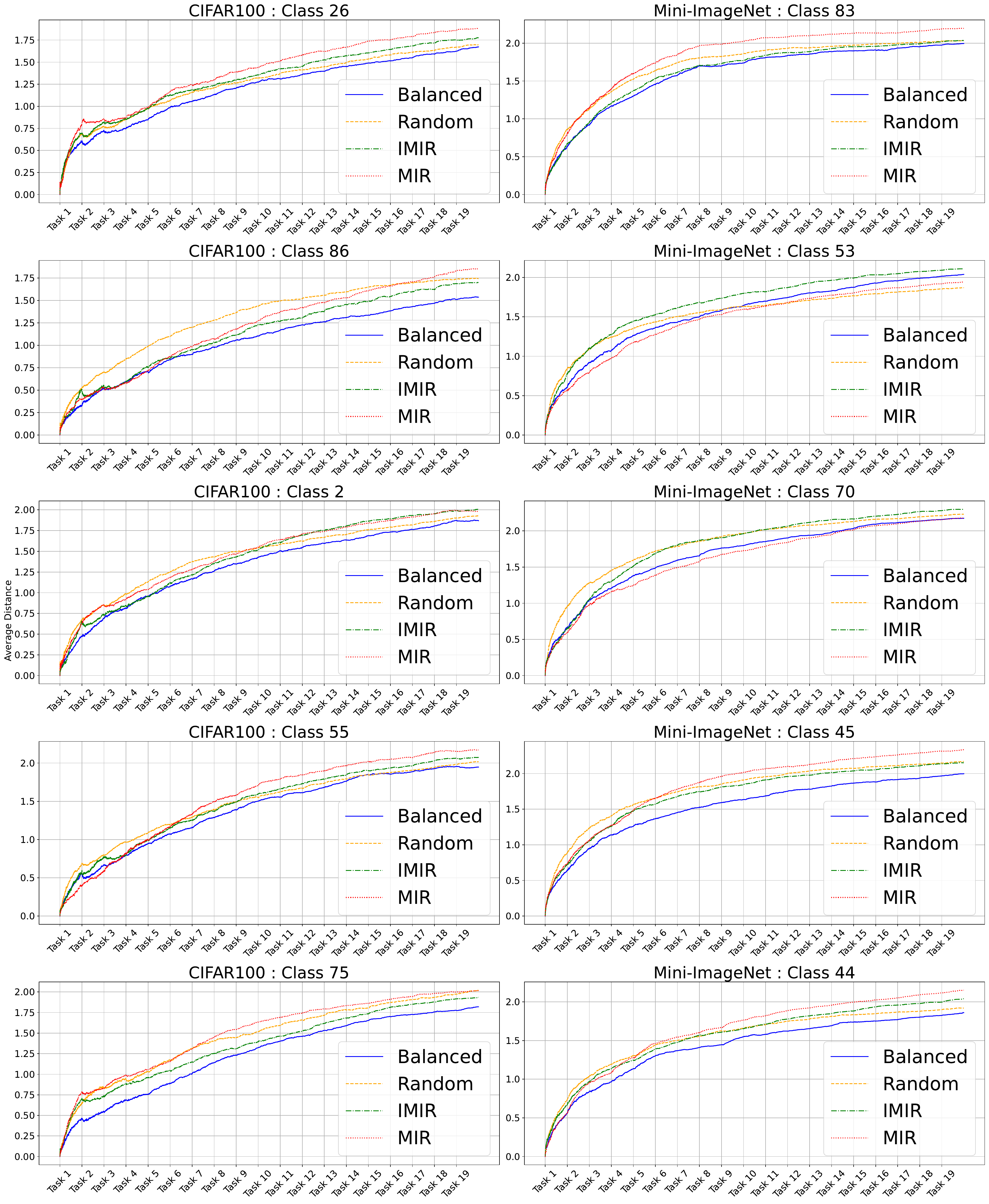}
\setlength{\abovecaptionskip}{-4mm} 
\caption{Proxies drift of 5 classes in task 0 on CIFAR100 and Mini-ImageNet.}
    \label{fig:proxies-drift-run0}
\vspace{-2mm}
\end{figure*}

\noindent \textbf{Class-based proxy drift.} In the main paper, we presented the average proxy drift across all classes within selected tasks. To gain deeper insights, we analyzed proxy drift at the level of individual classes. Fig.~\ref{fig:proxies-drift-run0} illustrates the proxy drift for each class in the first task (Task 0) on CIFAR100 and Mini-ImageNet. These two datasets were chosen as their proxy drift patterns are representative of the trends observed across the remaining benchmarks.

Two notable patterns emerge from the analysis. First, for all classes in the first task (particularly in CIFAR100), a sudden and significant proxy shift occurs after the introduction of the third task. This indicates that proxy drift escalates early in the continual learning process, with random retrieval failing entirely to mitigate it. Second, while targeted retrieval methods such as MIR and IMIR initially compensate for drift, their robustness diminishes as later tasks are introduced, typically becoming evident midway through the task sequence.

In contrast, the proposed balanced retrieval consistently demonstrates superior drift robustness, both in the early stages and as tasks progress. Moreover, its advantage over MIR and IMIR becomes more pronounced in the latter tasks. These findings highlight that the proposed balanced retrieval effectively reduces proxy drift, making it particularly suitable for long task sequences in continual learning scenarios.

\begin{figure*}[h!]
  \centering
      \includegraphics[width=1 \linewidth]{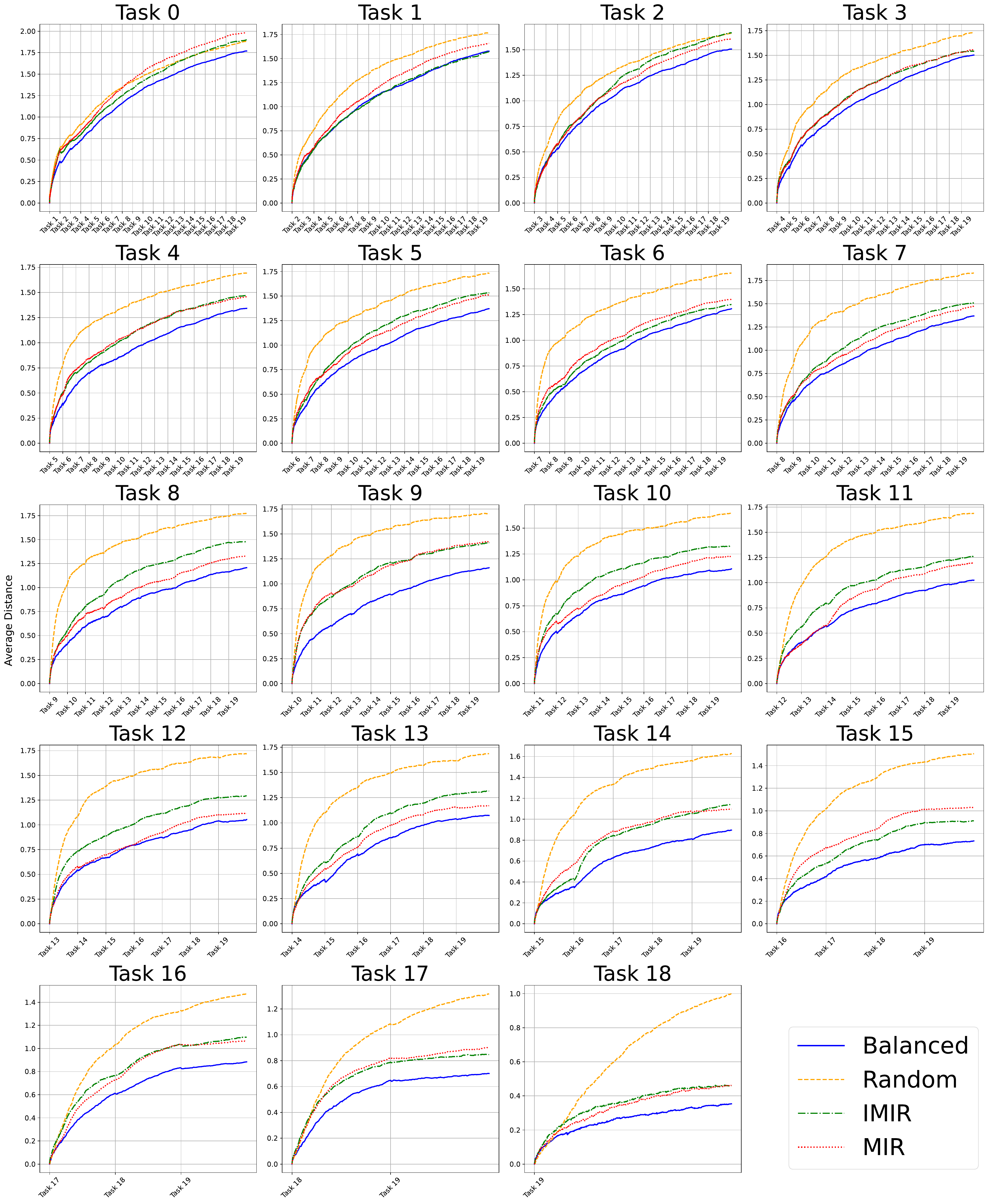}
\setlength{\abovecaptionskip}{-4mm} 
\caption{The proxy shift of classes in each task of the CIFAR100.}
    \label{fig:proxies-drift-task}
\vspace{-2mm}
\end{figure*}

\begin{figure*}[h!]
  \centering
      \includegraphics[width=1 \linewidth]{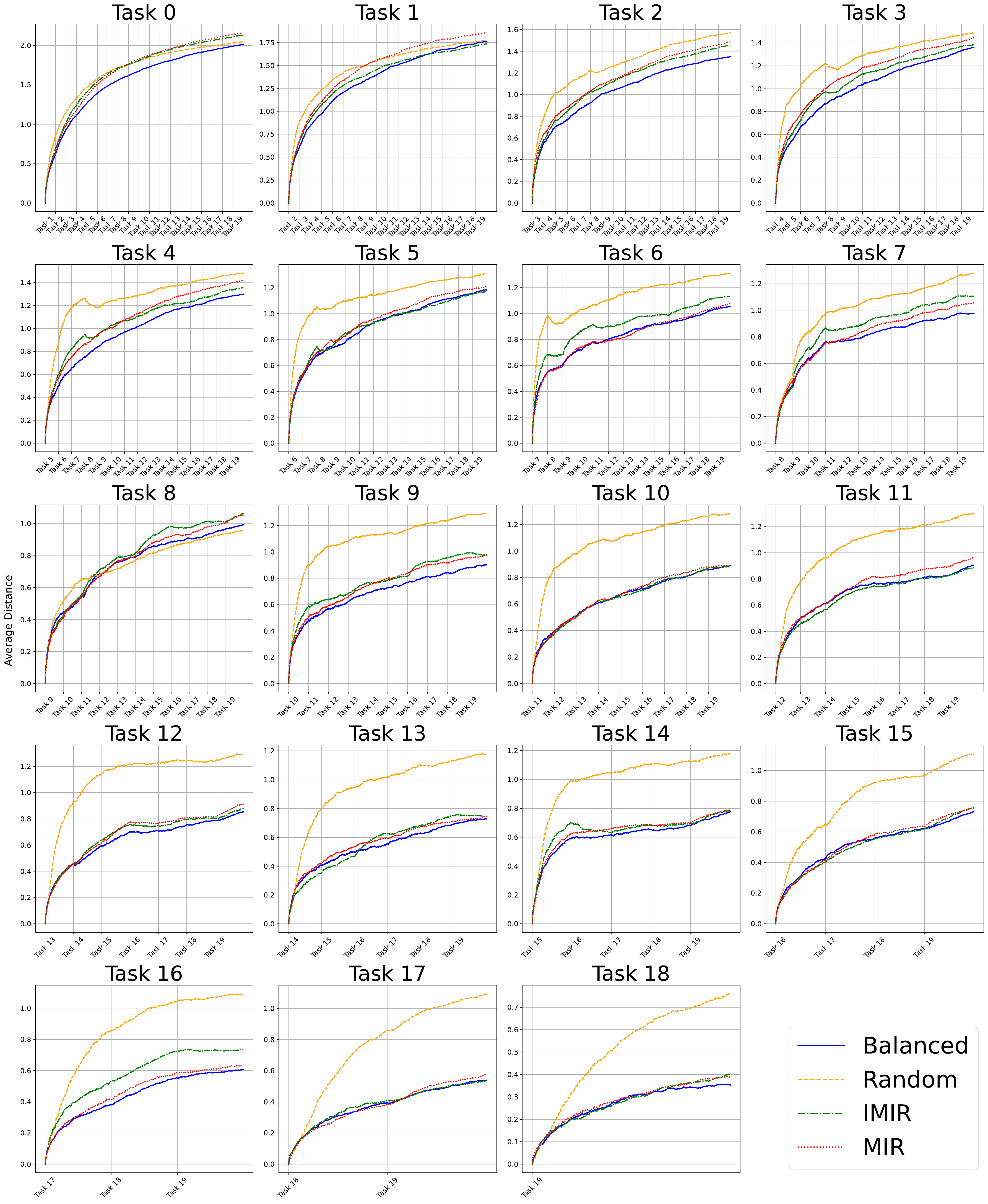}
\setlength{\abovecaptionskip}{-4mm} 
\caption{The proxy shift of classes in each task of the Mini-ImageNet.}
    \label{fig:proxies-drift-task2}
\vspace{-2mm}
\end{figure*}

\noindent \textbf{Task-based proxy drift.} Fig.~\ref{fig:proxies-drift-task} and~\ref{fig:proxies-drift-task2} present the average proxy drift per task (calculated across all classes within each task) for CIFAR100 and Mini-ImageNet. The results align with the trends reported in the main paper for selected tasks. Notably, random retrieval exhibits a marked increase in proxy drift during later tasks. In contrast, all targeted retrieval methods demonstrate improved robustness against drift in these later stages, with the proposed balanced retrieval consistently achieving the lowest drift magnitude. These findings underscore the critical role of targeted retrieval strategies in proxy-based continual learning, particularly for mitigating drift towards the end of long task streams.

\begin{figure*}[h!]
  \centering
      \includegraphics[width=1 \linewidth]{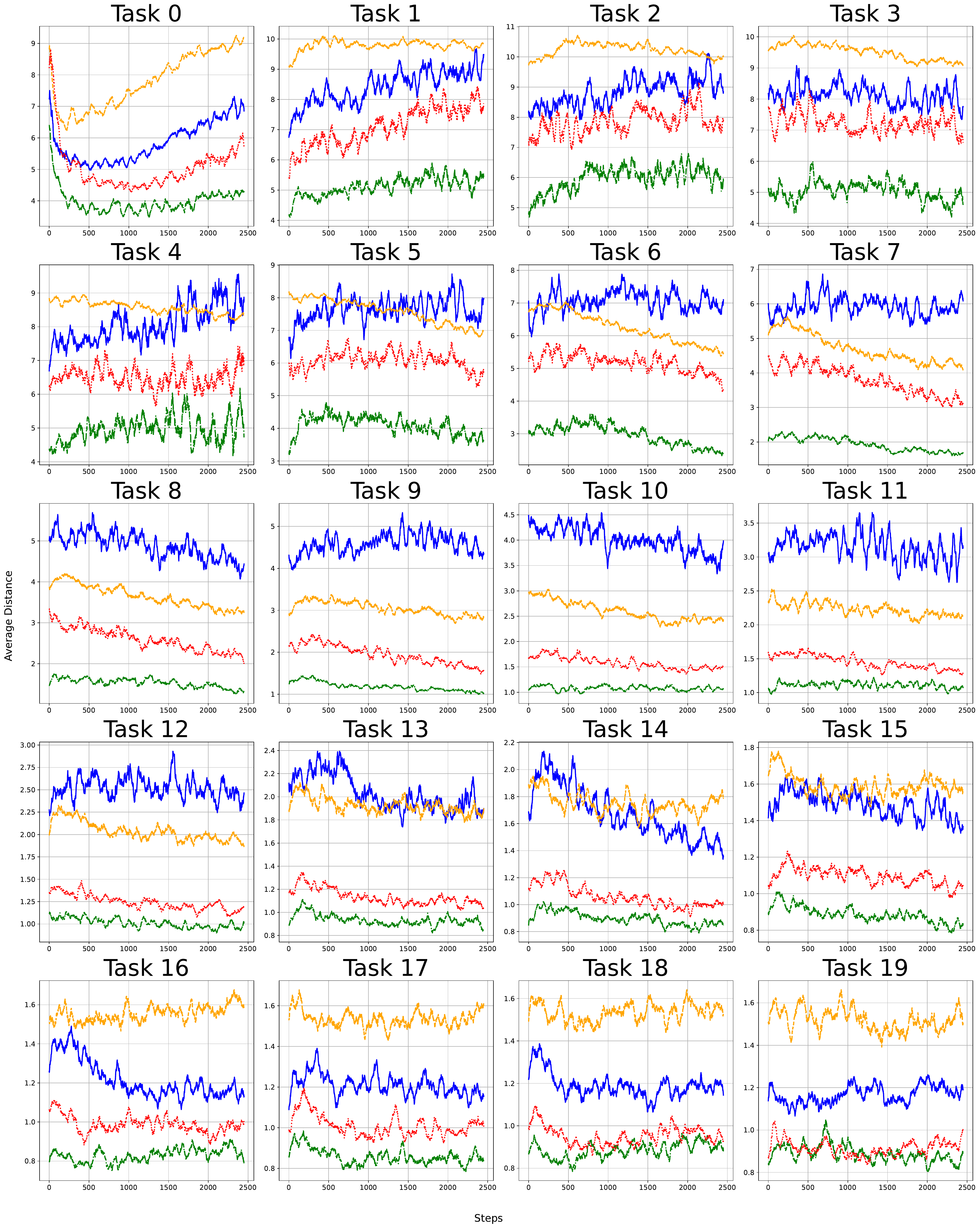}
\setlength{\abovecaptionskip}{-4mm} 
\caption{The average inner distance of retrieved samples during the training of all tasks on CIFAR100.The methods are represented by the following colors:
\colorbox{MatplotlibBlue}{\textcolor{white}{Balanced}},
\colorbox{MatplotlibOrange}{\textcolor{black}{Random}},
\colorbox{MatplotlibGreen}{\textcolor{white}{MIR}},
\colorbox{MatplotlibRed}{\textcolor{white}{IMIR}}.}
    \label{fig:inner-distance-run0}
\vspace{-2mm}
\end{figure*}

\begin{figure*}[h!]
  \centering
      \includegraphics[width=1 \linewidth]{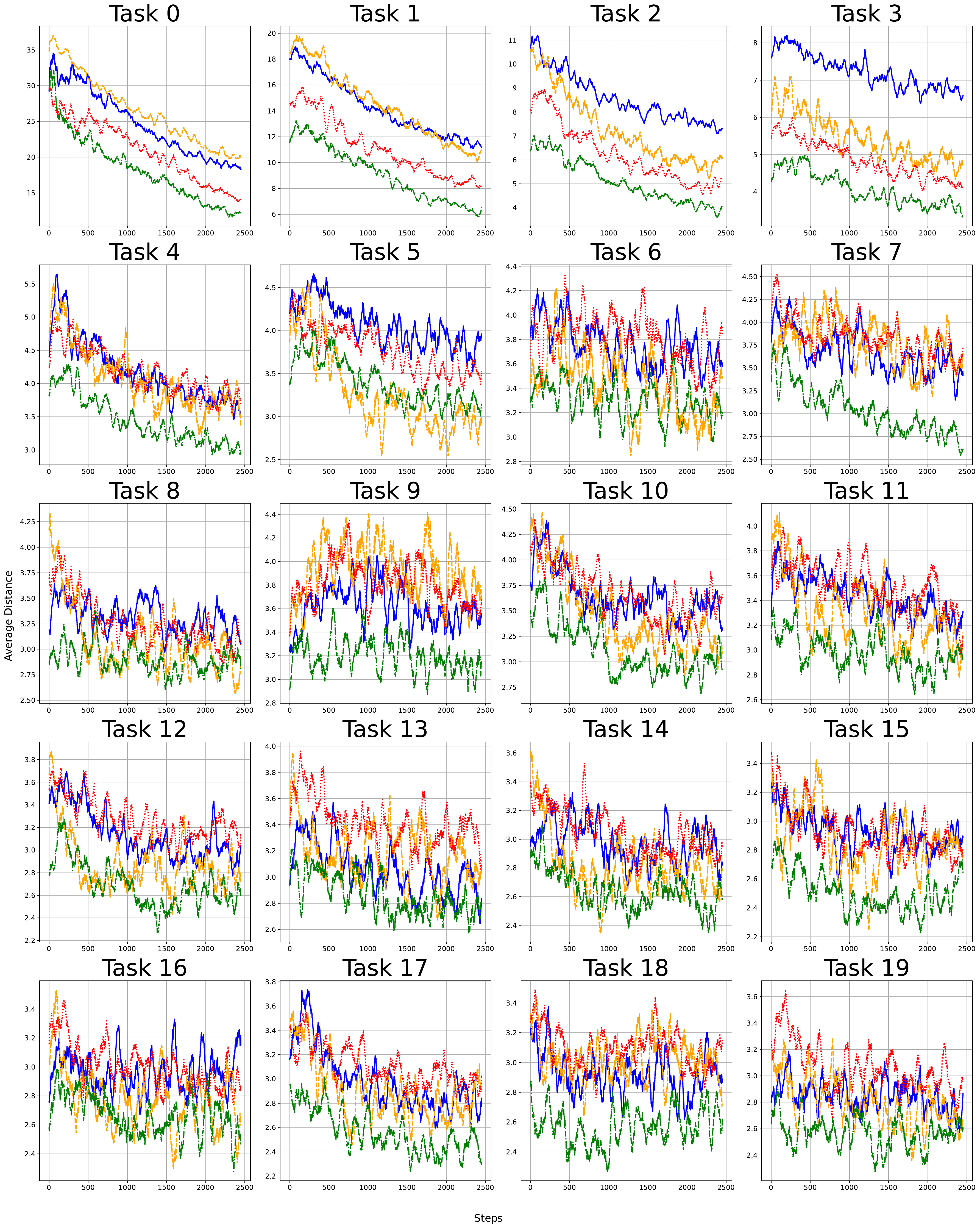}
\setlength{\abovecaptionskip}{-4mm} 
\caption{The average inner distance of retrieved samples during the training of all tasks on Mini-ImageNet.The methods are represented by the following colors:
\colorbox{MatplotlibBlue}{\textcolor{white}{Balanced}},
\colorbox{MatplotlibOrange}{\textcolor{black}{Random}},
\colorbox{MatplotlibGreen}{\textcolor{white}{MIR}},
\colorbox{MatplotlibRed}{\textcolor{white}{IMIR}}.}
    \label{fig:inner-distance-run0-mini}
\vspace{-2mm}
\end{figure*}

\begin{figure*}[h!]
  \centering
      \includegraphics[width=1 \linewidth]{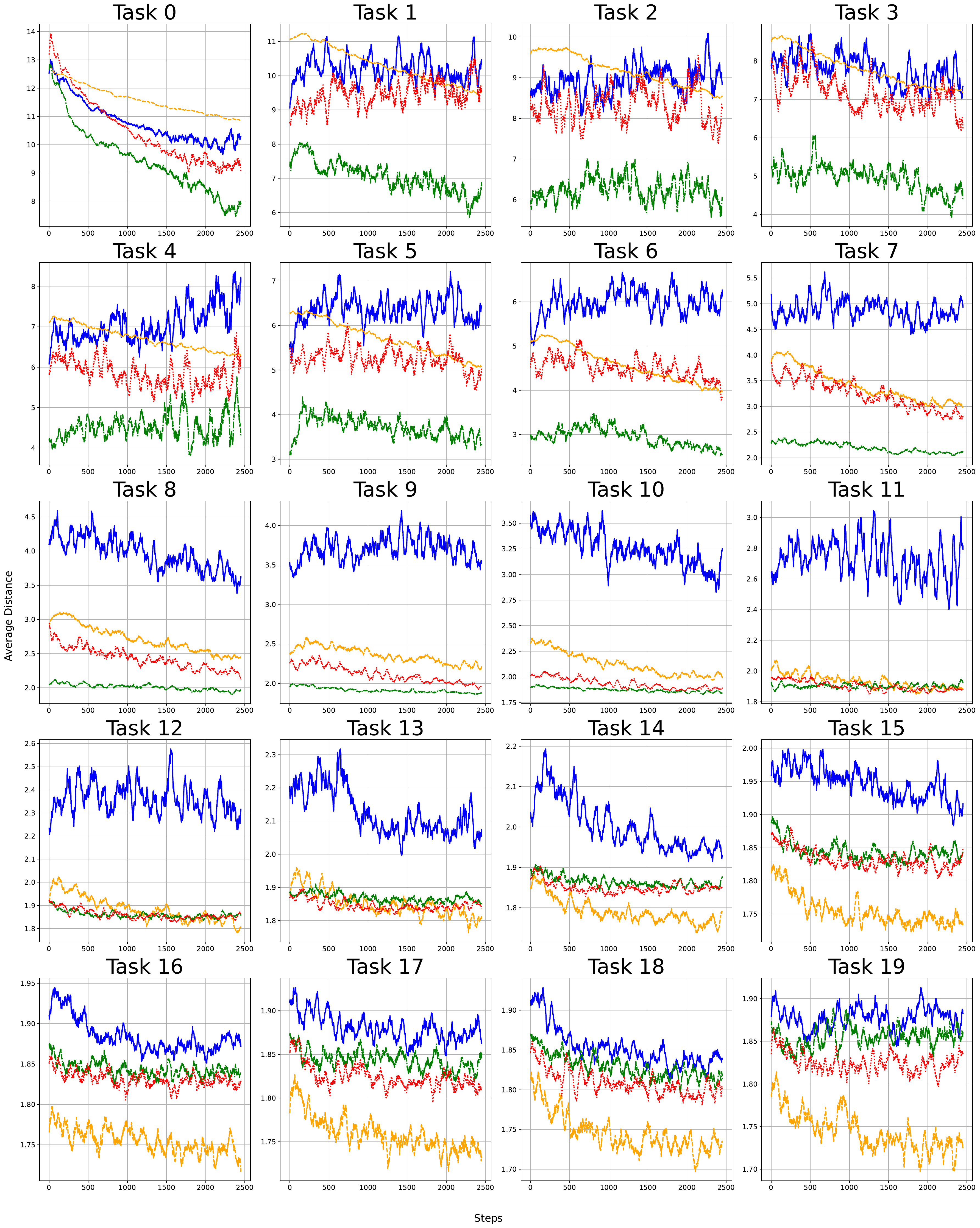}
\setlength{\abovecaptionskip}{-4mm} 
\caption{The average distance between the retrieved samples and their corresponding proxies during the training of all tasks on the CIFAR100.The methods are represented by the following colors:
\colorbox{MatplotlibBlue}{\textcolor{white}{Balanced}},
\colorbox{MatplotlibOrange}{\textcolor{black}{Random}},
\colorbox{MatplotlibGreen}{\textcolor{white}{MIR}},
\colorbox{MatplotlibRed}{\textcolor{white}{IMIR}}.}
\label{fig:distance-to-proxies-run0}
\vspace{-2mm}
\end{figure*}

\begin{figure*}[h!]
  \centering
      \includegraphics[width=1 \linewidth]{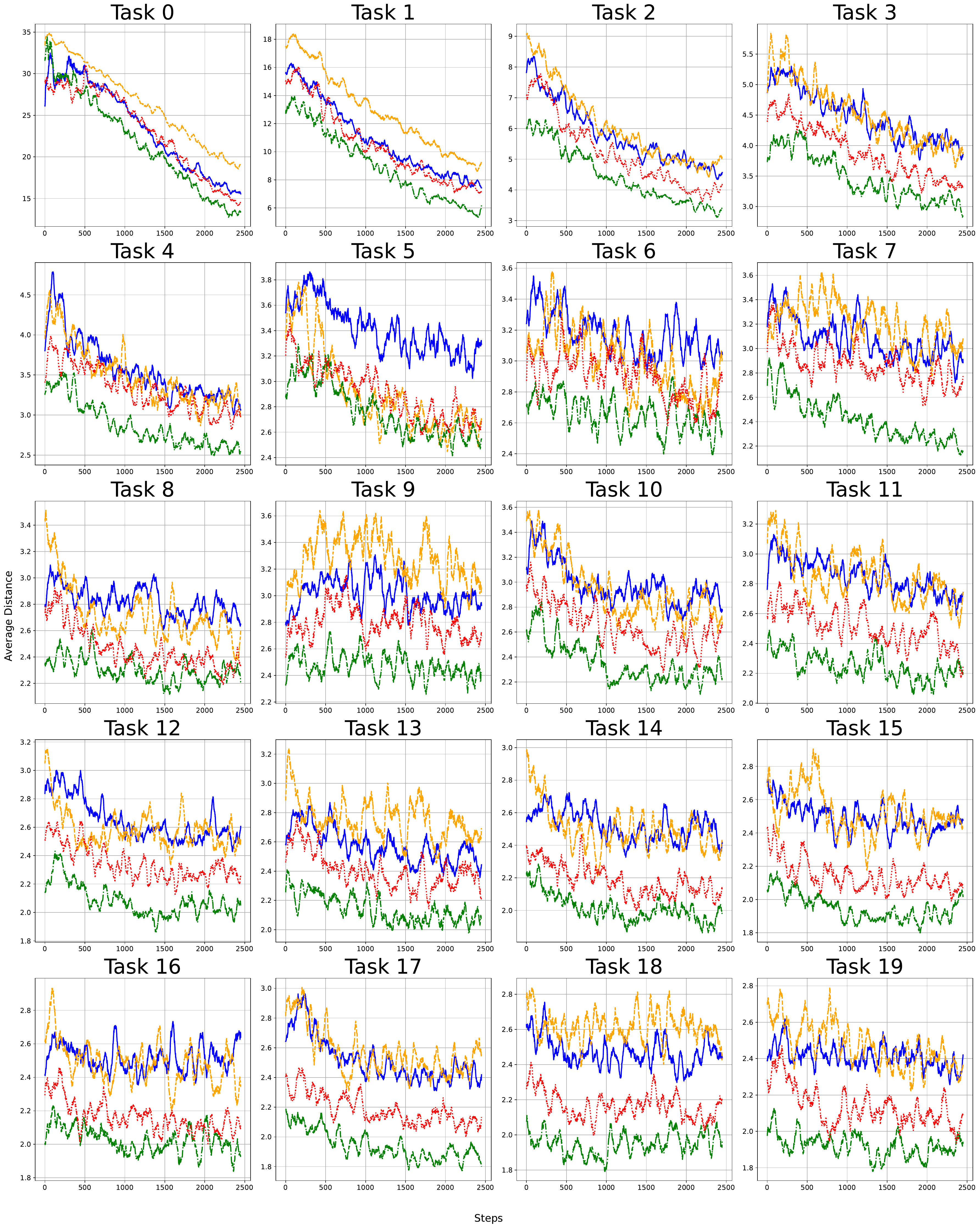}
\setlength{\abovecaptionskip}{-4mm} 
\caption{The average distance between the retrieved samples and their corresponding proxies during the training of all tasks in on the Mini-ImageNet.The methods are represented by the following colors:
\colorbox{MatplotlibBlue}{\textcolor{white}{Balanced}},
\colorbox{MatplotlibOrange}{\textcolor{black}{Random}},
\colorbox{MatplotlibGreen}{\textcolor{white}{MIR}},
\colorbox{MatplotlibRed}{\textcolor{white}{IMIR}}.}
    \label{fig:distance-to-proxies-run0_mini1}
\vspace{-2mm}
\end{figure*}

\noindent \textbf{Diversity of retrieved instances.} In the main paper, we presented selected results on the diversity of instanced retrieved by analyzed methods, measured as (i) inner distance of retrieved samples during the training; and (ii) average distance between the retrieved samples and their corresponding proxies. Fig.\ref{fig:inner-distance-run0} and \ref{fig:inner-distance-run0-mini} depict the inner distance of retrieved samples for each task in CIFAR 100 and Mini-ImageNet, while Fig.\ref{fig:distance-to-proxies-run0} and \ref{fig:distance-to-proxies-run0_mini1} depict the average distance between the retrieved samples and their corresponding proxies in the same setting. The observations across all tasks in these two datasets are consistent with the selected results presented in the main paper. The proposed balanced retrieval leads to increased diversity of retrieved instances compared to both MIR and IMIR, leading to achieving an trade-off between adaptation to new data and memorization of previous tasks. Additionally, when retrieval lacks diversity, proxies may collapse into suboptimal representations that fail to span the feature space adequately. Diverse instance retrieval prevents this collapse by ensuring the proxies remain robust and distributed.

\begin{figure*}[t]
  \centering
      \includegraphics[width=1 \linewidth]{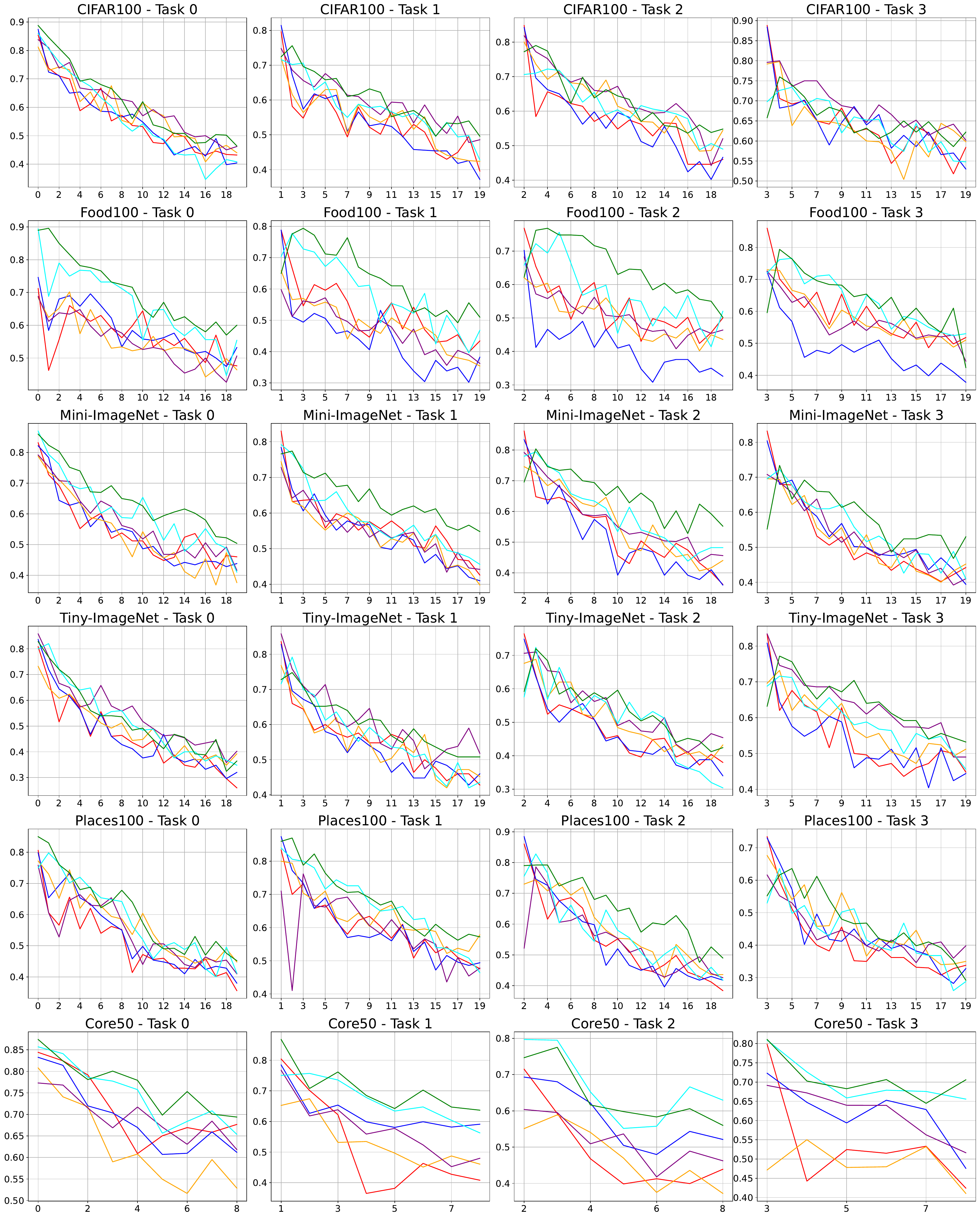}
\captionsetup{aboveskip=-0.5mm} 
\caption{The average accuracy for selected classes(task 0 to task 3) for different models after subsequent task batches.The methods are represented by the following colors:
        The methods are represented by the following colors:
        \colorbox{MatplotlibRed}{\textcolor{white}{CLSER}},
        \colorbox{MatplotlibBlue}{\textcolor{white}{CLSER+OURS}},
        \colorbox{MatplotlibOrange}{\textcolor{black}{ERACE}},
        \colorbox{MatplotlibPurple}{\textcolor{white}{ERACE+OURS}},
        \colorbox{MatplotlibCyan}{\textcolor{black}{PCR}},
        \colorbox{MatplotlibGreen}{\textcolor{white}{PCR+OURS}}.}
    \label{fig:class_0to3}
\vspace{-2mm}
\end{figure*}

\begin{figure*}[t]
  \centering
      \includegraphics[width=1 \linewidth]{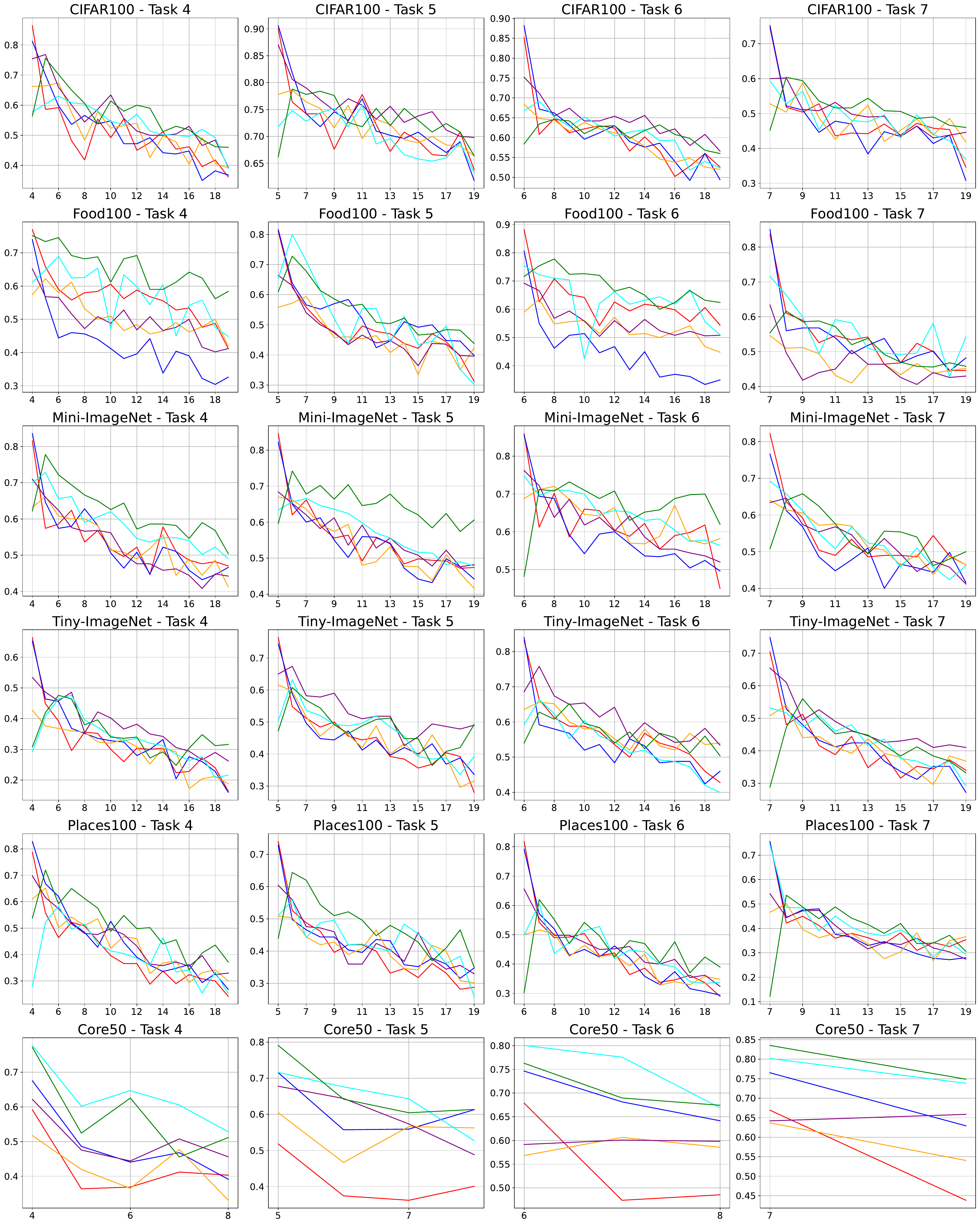}
\captionsetup{aboveskip=-0.5mm} 
\caption{The average accuracy for selected classes(task 4 to task 7) for different models after subsequent task batches.The methods are represented by the following colors:
        The methods are represented by the following colors:
        \colorbox{MatplotlibRed}{\textcolor{white}{CLSER}},
        \colorbox{MatplotlibBlue}{\textcolor{white}{CLSER+OURS}},
        \colorbox{MatplotlibOrange}{\textcolor{black}{ERACE}},
        \colorbox{MatplotlibPurple}{\textcolor{white}{ERACE+OURS}},
        \colorbox{MatplotlibCyan}{\textcolor{black}{PCR}},
        \colorbox{MatplotlibGreen}{\textcolor{white}{PCR+OURS}}.}
    \label{fig:class_4to7}
\vspace{-2mm}
\end{figure*}

\begin{figure*}[t]
  \centering
      \includegraphics[width=1 \linewidth]{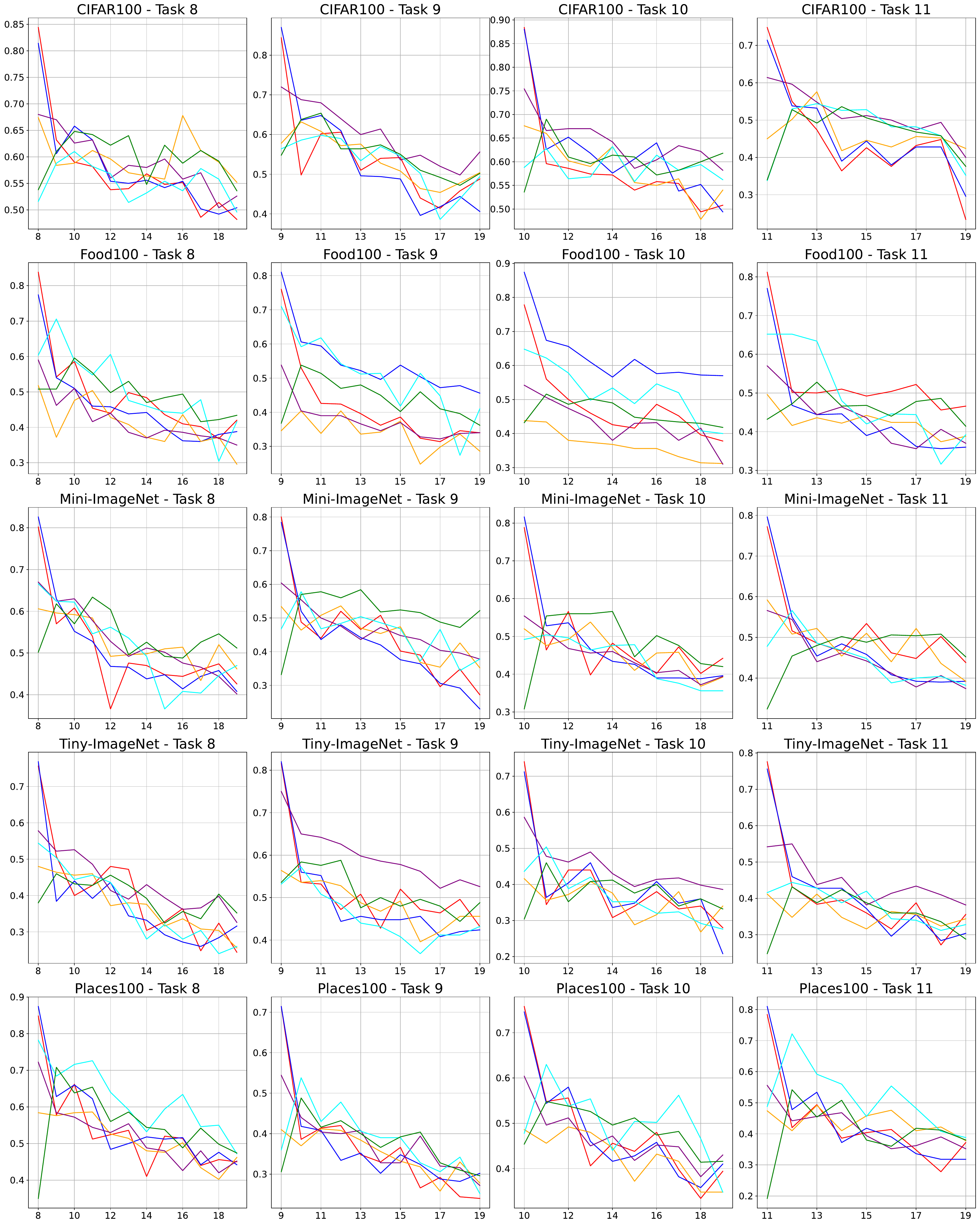}
\captionsetup{aboveskip=-0.5mm} 
\caption{The average accuracy for selected classes(task 8 to task 11) for different models after subsequent task batches.The methods are represented by the following colors:
        The methods are represented by the following colors:
        \colorbox{MatplotlibRed}{\textcolor{white}{CLSER}},
        \colorbox{MatplotlibBlue}{\textcolor{white}{CLSER+OURS}},
        \colorbox{MatplotlibOrange}{\textcolor{black}{ERACE}},
        \colorbox{MatplotlibPurple}{\textcolor{white}{ERACE+OURS}},
        \colorbox{MatplotlibCyan}{\textcolor{black}{PCR}},
        \colorbox{MatplotlibGreen}{\textcolor{white}{PCR+OURS}}.}
    \label{fig:class_8to11}
\vspace{-2mm}
\end{figure*}

\begin{figure*}[t]
  \centering
      \includegraphics[width=1 \linewidth]{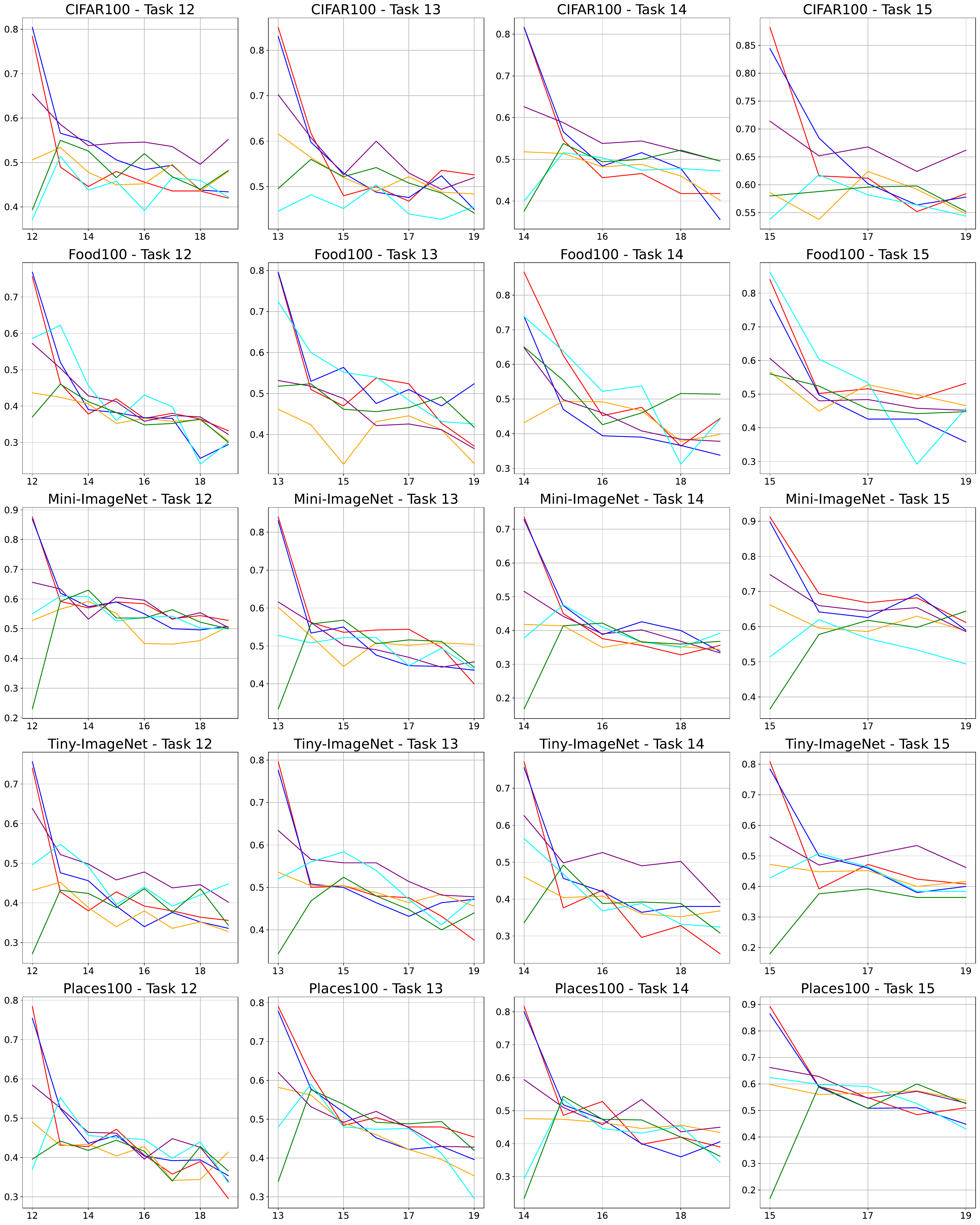}
\captionsetup{aboveskip=-0.5mm} 
\caption{The average accuracy for selected classes(task 12 to task 15) for different models after subsequent task batches.The methods are represented by the following colors:
        The methods are represented by the following colors:
        \colorbox{MatplotlibRed}{\textcolor{white}{CLSER}},
        \colorbox{MatplotlibBlue}{\textcolor{white}{CLSER+OURS}},
        \colorbox{MatplotlibOrange}{\textcolor{black}{ERACE}},
        \colorbox{MatplotlibPurple}{\textcolor{white}{ERACE+OURS}},
        \colorbox{MatplotlibCyan}{\textcolor{black}{PCR}},
        \colorbox{MatplotlibGreen}{\textcolor{white}{PCR+OURS}}.}
    \label{fig:class_12to15}
\vspace{-2mm}
\end{figure*}

\begin{figure*}[t]
  \centering
      \includegraphics[width=1 \linewidth]{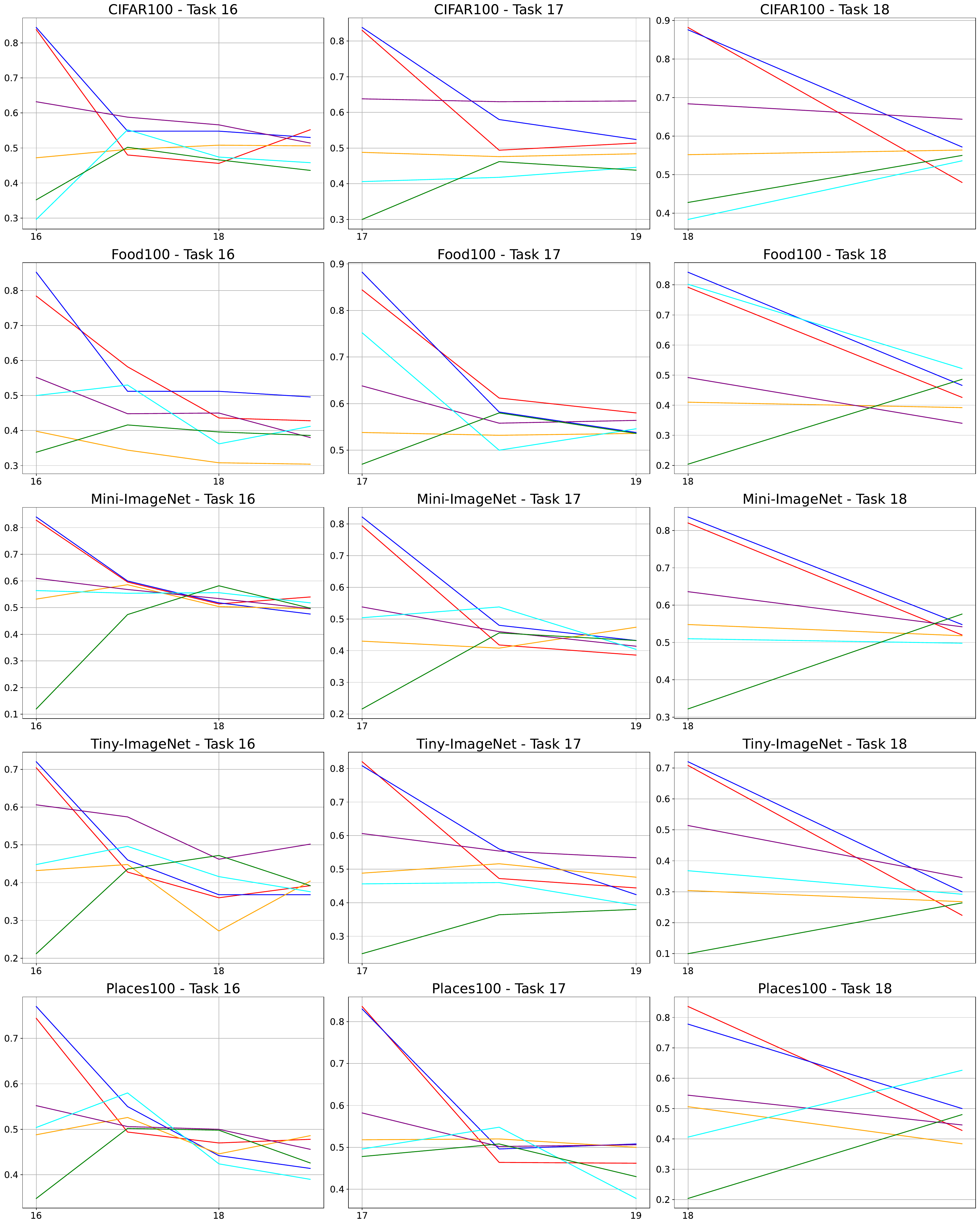}
\captionsetup{aboveskip=-0.5mm} 
\caption{The average accuracy for selected classes(task 16 to task 18) for different models after subsequent task batches.The methods are represented by the following colors:
        The methods are represented by the following colors:
        \colorbox{MatplotlibRed}{\textcolor{white}{CLSER}},
        \colorbox{MatplotlibBlue}{\textcolor{white}{CLSER+OURS}},
        \colorbox{MatplotlibOrange}{\textcolor{black}{ERACE}},
        \colorbox{MatplotlibPurple}{\textcolor{white}{ERACE+OURS}},
        \colorbox{MatplotlibCyan}{\textcolor{black}{PCR}},
        \colorbox{MatplotlibGreen}{\textcolor{white}{PCR+OURS}}.}
    \label{fig:class_16to18}
\vspace{-2mm}
\end{figure*}

\noindent \textbf{Task accuracy of selected classifiers.} Fig.~\ref{fig:class_0to3} --~\ref{fig:class_16to18} present detailed per-task accuracies for all six continual learning benchmarks. These plots provide critical insights into the performance of the proposed instance retrieval method across task sequences. Specifically, they illustrate how balanced retrieval, when integrated with PCR, CLS-ER, and ER-ACE, improves the trade-off between maintaining performance on previously learned tasks and adapting to new ones. The visualized trends highlight the method's ability to mitigate catastrophic forgetting while ensuring effective knowledge transfer to emerging tasks, demonstrating a robust balance between stability and plasticity.

\newpage

\end{document}